\def\year{2022}\relax
\documentclass[letterpaper]{article} 
\usepackage{arxiv}  
\usepackage{times}  
\usepackage{helvet}  
\usepackage{courier}  
\usepackage[hyphens]{url}  
\usepackage{graphicx} 
\usepackage{natbib}  
\usepackage{caption} 
\usepackage{enumitem}
\DeclareCaptionStyle{ruled}{labelfont=normalfont,labelsep=colon,strut=off} 
\usepackage{algorithm}
\usepackage{algorithmic}
\usepackage{subfigure}
\usepackage{amsmath}
\usepackage{booktabs}
\usepackage{makecell}
\usepackage{diagbox}
\usepackage{enumerate}
\usepackage{multirow}
\usepackage{amssymb}
\usepackage{xcolor}
\usepackage{newfloat}
\usepackage{listings}
\floatstyle{ruled}
\newfloat{listing}{tb}{lst}{}
\floatname{listing}{Listing}
\title{Task Guided Representation Learning using Compositional Models for Zero-shot Domain Adaptation}
\author {
    Shuang Liu,\textsuperscript{\rm 1}
    Mete Ozay \textsuperscript{\rm 2}
}
\affiliations {
    \textsuperscript{\rm 1} RIKEN Center for AIP\\
}

\usepackage{bibentry}

\begin{document}

\maketitle

\begin{abstract}
Zero-shot domain adaptation (ZDA) methods aim to transfer knowledge about a task learned in a source domain to a target domain, while data from target domain are not available. In this work, we address learning feature representations which are invariant to and shared among different domains considering task characteristics for ZDA. To this end, we propose a method for task-guided ZDA (TG-ZDA) which employs multi-branch deep neural networks to learn feature representations exploiting their domain invariance and shareability properties. The proposed TG-ZDA models can be trained end-to-end without requiring synthetic tasks and data generated from estimated representations of target domains. The proposed TG-ZDA has been examined using benchmark ZDA tasks on image classification datasets. Experimental results show that our proposed TG-ZDA outperforms state-of-the-art ZDA methods for different domains and tasks.
\end{abstract}

\section{Introduction}
\label{sec:intro}
In real-world applications, distribution of training data of a machine learning algorithm obtained from source domain $\mathcal{D}_{sr}$ may diverge from that of its test data obtained from target domain $\mathcal{D}_{t}$. This problem is called domain shift. Machine learning models trained under domain shift suffer from a significant performance drop. For example, a model trained for detecting vehicles using a dataset collected in sunny days (e.g. from a source domain $\mathcal{D}_{sr}$) performs well for inference on another dataset of vehicles collected in sunny days (e.g. from the same domain $\mathcal{D}_{sr}$). However, the trained model may perform poorly on a dataset of vehicles collected in rainy days (e.g. from a target domain $\mathcal{D}_{t}$). 
This problem becomes acute for deep neural networks (DNNs) which require massive amount of data for training. 

A solution to this problem is fine-tuning models pre-trained on $\mathcal{D}_{sr}$ with data collected from $\mathcal{D}_{t}$. However, this solution demands considerable amount of labeled training data from $\mathcal{D}_{t}$, which may not be   available in many applications, e.g. medical informatics. To overcome this problem, \textit{Domain adaptation (DA)} methods aim to train a model on a dataset from $\mathcal{D}_{sr}$, so that the trained model can perform well on a test dataset from $\mathcal{D}_{t}$. DA methods which train models using unlabelled data from $\mathcal{D}_{t}$ are called unsupervised domain adaptation (UDA) methods \cite{bousmalis2017unsupervised,Lee2019SlicedWD,Chen2020HoMMHM}. UDA methods require large amount of unlabelled data from $\mathcal{D}_{t}$ to train models. However, this assumption may not be valid in various real-world applications. For instance, access to $\mathcal{D}_{t}$ may be limited or $\mathcal{D}_{t}$ may be updated dynamically as in lifelong learning and autonomous driving.
To address this problem for training models without using labels and unlabelled data from $\mathcal{D}_{t}$, zero-shot domain adaptation (ZDA) methods were proposed \cite{Peng2018ZeroShotDD,Wang2019ConditionalCG}.


ZDDA~\cite{zddaTpami} is a DNN based method proposed for ZDA. ZDDA models are trained in three phases across four modules, therefore, they are inconvenient to be deployed in various applications. CoCoGAN~\cite{LATZDA} and its variations~\cite{wang2020adversarial,DSPZDA} extended CoGAN~\cite{Liu2016CoupledGA} to synthesize samples in missing  domains for solving the ZDA problem. However, they cannot guarantee that their proposed GAN learns the real distribution of the missing domains. HGnet~\cite{Xia2020HGNetHG} 
aims to learn domain invariant features, but does not incorporate feature shareability among domains and across tasks, which is demonstrated to be beneficial for ZDA.

\begin{figure}[!t]
\centering
\includegraphics[width=0.52\textwidth]{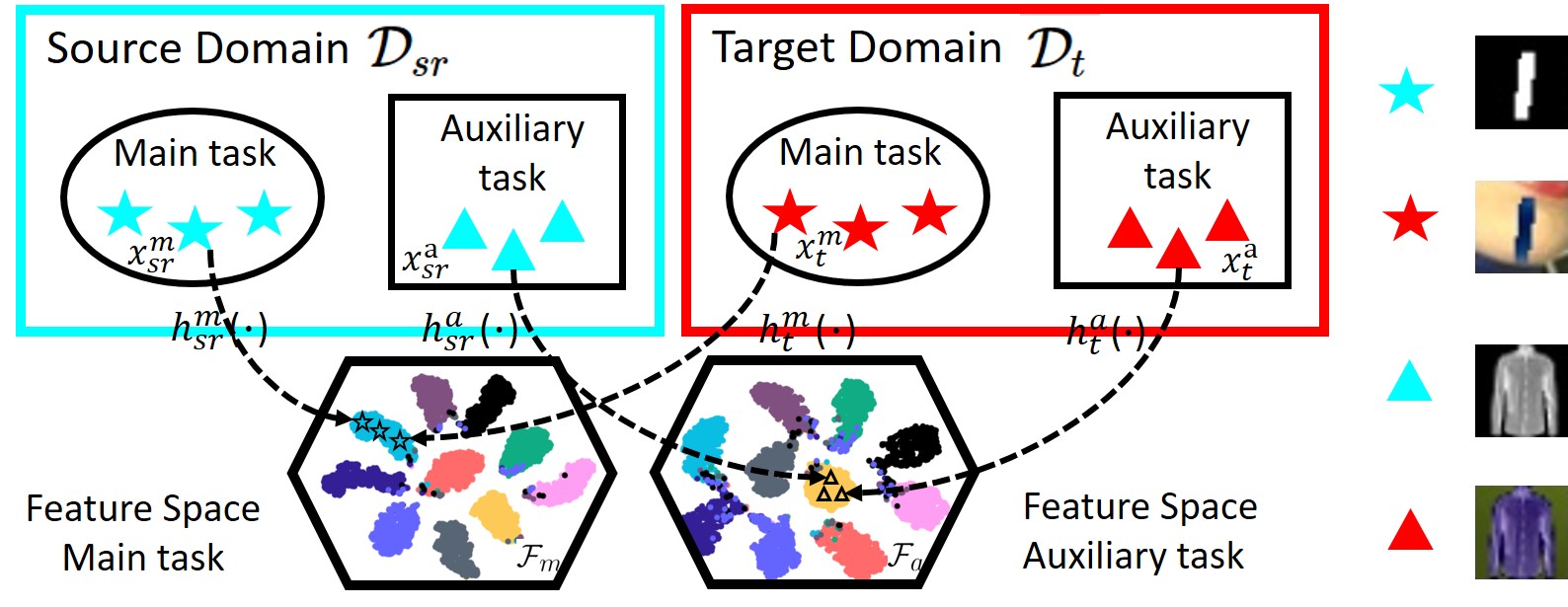}
\caption{An illustration of our proposed ZDA problem and model. Please see the text for notation.}
\label{fig:problem1}
\end{figure}

We address the ZDA problem (Figure~\ref{fig:problem1}) employing invariance \cite{zhao19a} and shareability \cite{liang2020shot} of hypotheses of feature representations in a compositional framework motivated by the recent advances in DA theory \cite{creager21environment,KamathTSS21,0080WZLKD021}, and observations in deep learning theory~\cite{lecun2015deeplearning,Saxe11537,Bau30071,Lampinen32970,Sejnowski30033,Poggio30039}. Our contributions are summarized as follows:

$\bullet$ 
     We propose a ZDA method called Task-guided Multi-branch Zero-shot Domain Adaptation (TG-ZDA) depicted in Figure~\ref{fig:arc},  which takes advantage of invariance and shareability of learned features by the guidance of auxiliary tasks (Figure~\ref{fig:problem1}). In TG-ZDA, we first estimate hypotheses of domain invariant and shareable features. To transfer knowledge among domains governing auxiliary tasks, we learn task-aware hypotheses. Then, we integrate them in composite hypotheses by employing a novel end-to-end trainable DNN, since hierarchical compositional functions can be approximated by DNNs without incurring the curse of dimensionality \cite{Poggio30039}. TG-ZDA does not require synthesizing tasks and artificial data from target domains for training models unlike state-of-the-art ZDA methods such as CoCoGAN~\cite{LATZDA}. 
    
    $\bullet$  The proposed TG-ZDA has been analyzed on multiple ZDA tasks in comparison with state-of-the-art methods. In the analyses, TG-ZDA outperforms the state-of-the-art methods by 8.2\% on average over 50 classification tasks. 
    
    $\bullet$  In the analyses, we first visually explore learned domain invariant, shareable and composite feature representations. Then, we compute the Fowlkes-Mallows index (FMI) and normalized mutual information (NMI) of clusters of these feature representations for their statistical analysis.


\section{Related Work}
DA methods are categorized as unsupervised, few-shot and zero-shot DA according to statistics of data obtained from a target domain $\mathcal{D}_{t}$ which can be accessed during training. 


\textbf{Unsupervised domain adaptation (UDA):} UDA methods employ large amount of unlabelled data belonging to $\mathcal{D}_{t}$. Distance-based UDA methods use multi-stream DNNs to align features obtained from different streams by minimizing their domain discrepancy at  certain layers for training models. A widely used distance metric proposed for measuring domain discrepancy is Maximum Mean Discrepancy (MMD) \cite{borgwardt2006integrating,tzeng2014deep}. 
Correlation alignment \cite{sun2015return} was proposed to minimize domain shift through aligning distributions with linear or nonlinear \cite{sun2016deep} transformations.  
Adversarial DA models  utilize a  discriminator aiming to discriminate features among domains while a feature extractor aims to produce indistinguishable features from different domains \cite{ganin2016domain,tzeng2017adversarial,xu2019adversarial,Tang2020DiscriminativeAD,Liu2019TransferableAT,Luo2019TakingAC}.
Other approaches aim to overcome DA problems in pixel spaces \cite{bousmalis2017unsupervised,huang2018auggan,generativedocking}.

\textbf{Few-shot domain adaptation (FDA):} UDA methods demand large amount of unlabeled training data from $\mathcal{D}_{t}$ which may not be available in practice. FDA methods, which have been developed to solve this problem, assume that only a few labeled samples from $\mathcal{D}_{t}$ are available during training.  \citet{Wang2019FewShotAF} adapted Faster-RCNN to target domains using a few labeled samples by introducing a pairing mechanism, designing new DA modules and imposing proper regularization. \citet{xu2019d} proposed stochastic embedding techniques to learn domain invariant features with a few labeled samples. Four groups of pairs were created by \citet{motiian2017few} to alleviate the lack of training data. 

\textbf{Zero-shot domain adaptation (ZDA):} 
ZDA methods do not have access to data from $\mathcal{D}_{t}$, which is a case commonly observed in many real-world applications  due to finite budget, limited development time or unavailability of $\mathcal{D}_{t}$. When data from $\mathcal{D}_{t}$ are not available for training models, the models use auxiliary information provided by irrelevant tasks to assist ZDA models. An assumption of these models is that task-irrelevant and task-relevant datasets contain information on source and target domains. ZDDA \cite{Peng2018ZeroShotDD} uses task-relevant data and task-irrelevant dual domain pairs as inputs. Exchanging target domain knowledge between relevant and irrelevant tasks is achieved via alternating training. 
Some works deployed generative DNNs to approximate the distribution of data in target domain \cite{Wang2019ConditionalCG,wang2020adversarial,DSPZDA,LATZDA}. Conditional coupled GANs (CoCoGANs) \cite{LATZDA} extended coupled GANs (CoGANs) \cite{Liu2016CoupledGA} with a binary conditioning signal for ZDA. It proposed a representation alignment method to generate target domain images. \citet{DSPZDA} introduced dual CoGANs consisting of a task-irrelevant and a task of interest CoGAN, for ZDA. The former is trained normally due to availability of data from dual domains. The latter is trained to synthesize task of interest samples in $\mathcal{D}_{t}$ by employing domain shift preservation constraints and co-training classifiers with supervisory signals. Although generative DNNs have shown superior results in generating images, bias still exists between generated and real distribution. The synthesized datasets have to be screened manually for training classifiers.  

\section{Method}
\label{sec:method}
\begin{figure*}[!h]
\centering
\includegraphics[width=\textwidth]{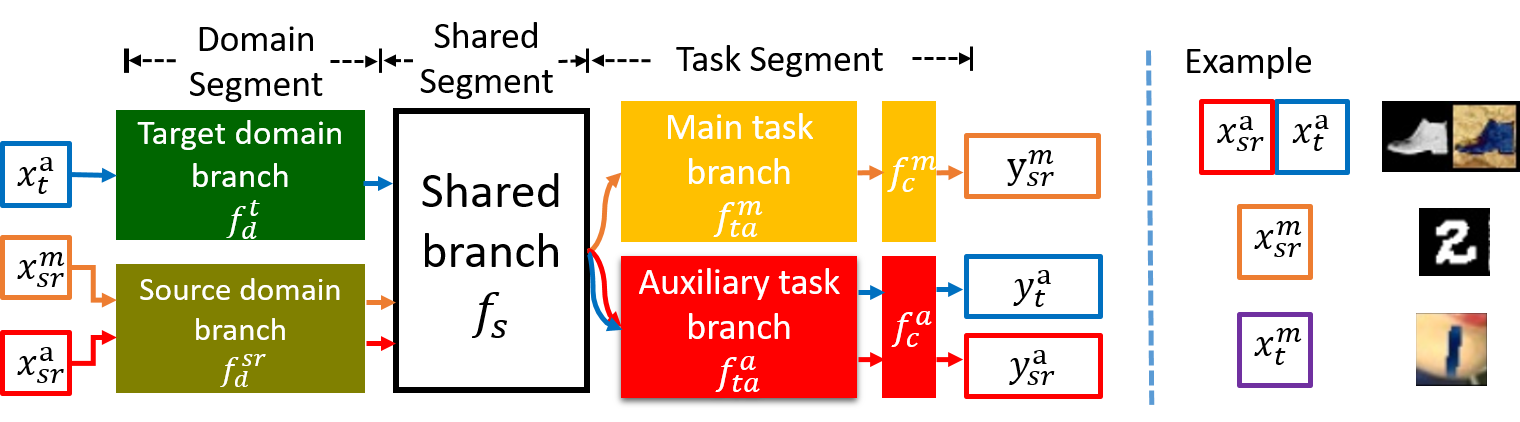}
\vspace{-0.4cm}
\caption{The architecture of the DNN of the TG-ZDA (left) and an input example (right).}
\label{fig:arc}

\end{figure*}

In this section,  we first provide our problem definition (Figure~\ref{fig:problem1}) and related notation. Then, our proposed TG-ZDA (Figure~\ref{fig:arc}) and the employed loss functions are introduced.

\begin{figure}[!tb]
\centering
\includegraphics[width=0.475\textwidth]{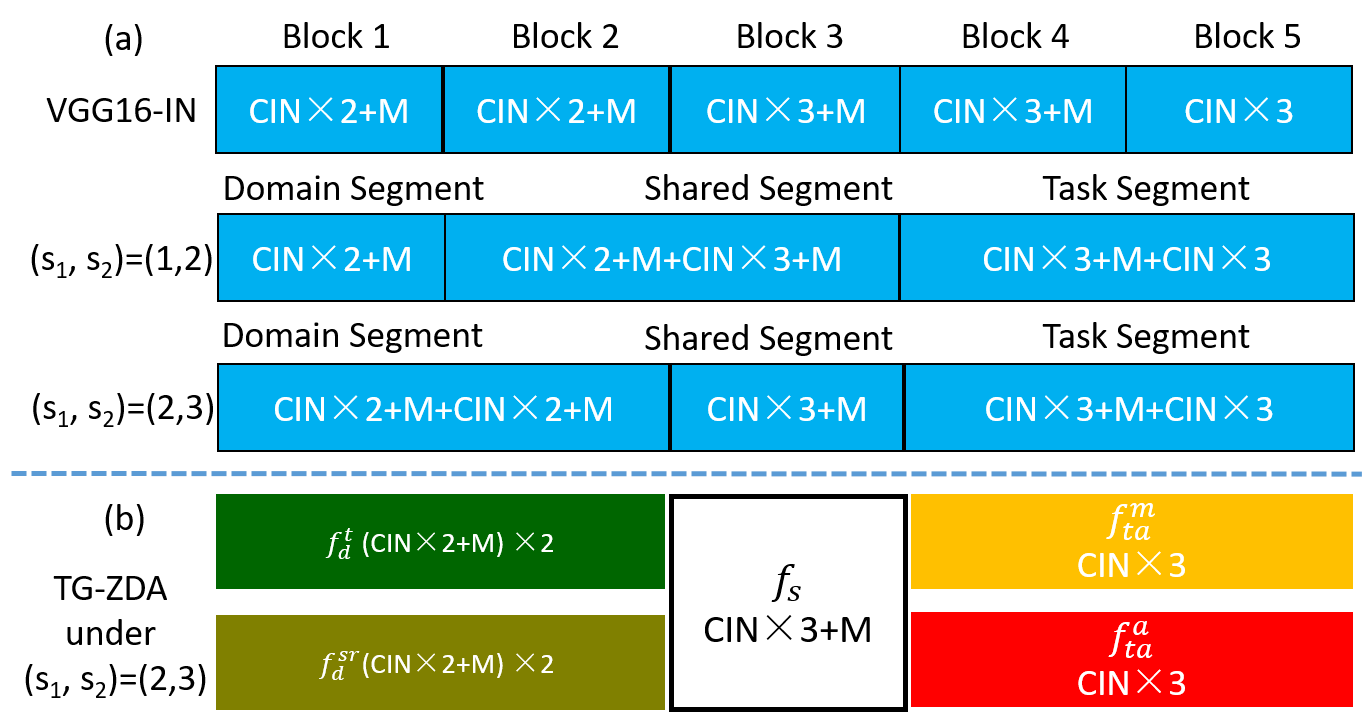}
\caption{An illustration of different group settings. CIN stands for a convolution layer followed by an instance normalization layer. M stands for a maxpooling layer.}
\label{fig:groupsetting2}
\end{figure}

\textbf{Problem Definition}: For ZDA, previous works \cite{Peng2018ZeroShotDD,Wang2019ConditionalCG} propose using two tasks; (i) a main task (a.k.a. relevant task), and (ii) an auxiliary task (a.k.a. irrelevant task). 
In the ZDA problem (Figure \ref{fig:problem1}), we use the main and auxiliary tasks to extract information from various domains:
\begin{itemize}
    \item \textbf{Main task} accesses the data $(x_{sr}^{m}, y_{sr}^{m}) \in \mathcal{X}_{sr}^{m} \times \mathcal{Y}_{sr}^{m}$ and $(x_{t}^{m}, y_{t}^{m}) \in \mathcal{X}_{t}^{m} \times \mathcal{Y}_{t}^{m}$ sampled from the domains $\mathcal{D}_{sr}$ and $\mathcal{D}_{t}$, respectively.

\item  \textbf{Auxiliary task} access the data ${(x_{sr}^{a}, y_{sr}^{a}) \in \mathcal{X}_{sr}^{a} \times \mathcal{Y}_{sr}^{a}}$ and $(x_{t}^{a},y_{t}^{a}) \in \mathcal{X}_{t}^{a} \times \mathcal{Y}_{t}^{a}$ sampled from the domains $\mathcal{D}_{sr}$ and $\mathcal{D}_{t}$, respectively. 
\end{itemize}
Datasets obtained from different domains and for different tasks are mapped to a feature space $\mathcal{F} = \mathcal{F}_{m} \cup \mathcal{F}_a$ by different hypotheses as depicted in Figure \ref{fig:problem1}, where $\mathcal{F}_{m}$ and $\mathcal{F}_{a}$ are feature spaces of main and auxiliary tasks, respectively. 
In the ZDA problem, we assume that we are given ${\mathcal{X} = \mathcal{X}_{sr}^{m} \cup \mathcal{X}_{sr}^{a} \cup \mathcal{X}_{t}^{a}}$ and ${\mathcal{Y} = \mathcal{Y}_{sr}^{m} \cup \mathcal{Y}_{sr}^{a} \cup \mathcal{Y}_{t}^{a}}$ in the training phase. Then, the goal of representation learning methods for ZDA is to estimate a composition ${h_c \circ h: x \in \mathcal{X} \mapsto y \in \mathcal{Y}}$ of feature representation hypotheses ($h \in \mathcal{H}: x \in \mathcal{X} \mapsto h(x) \in \mathcal{F}$) and classification hypotheses (${h_c \in \mathcal{H}: h(x) \in \mathcal{F} \mapsto y \in \mathcal{Y}}$),
such that the target risk $\mathbb{E}_{(x,y)\sim \mathcal{D}_t} l(f(x;{\mathcal{W}}), y)$ measured by a loss function $l(\cdot, \cdot)$ can be minimized by  functions $f(\cdot,{\mathcal{W}})$ approximating $h_c \circ h$ using DNNs.
That is, ZDA methods aim first to train a model using $\mathcal{X}$. Then, the trained model is tested on $\mathcal{X}_{t}^{m}$ belonging to the target domain $\mathcal{D}_t$. To identify structure of $f(\cdot,{\mathcal{W}})$ by DNNs and optimize its parameters $\mathcal{W}$, we propose a compositional framework called TG-ZDA depicted in Figure~\ref{fig:arc} and \ref{fig:groupsetting2}, and described next.

\vspace{0.05cm}
 \noindent
 \textbf{Estimating Composite Hypotheses in TG-ZDA}
 
As shown in Figure \ref{fig:groupsetting2}, data belonging to different domains and associated with different tasks are mapped to a domain independent feature space $\mathcal{F}$ by  
\begin{equation}
h_{\alpha}^{\beta} \in \mathcal{H}_{\alpha}^{\beta}: x_{\alpha}^{\beta} \in \mathcal{X}_{\alpha}^{\beta} \mapsto h_{\alpha}^{\beta}(x_{\alpha}^{\beta}) \in \mathcal{F}_{\beta},
\label{eq:hypothes}
\end{equation}
for all $\alpha \in \{sr, t\}, \beta \in \{m, a\}.$
Since the dataset $\mathcal{X}_{t}^{m}$ belonging to the target domain $\mathcal{D}_{t}$ is not available in the training phase, estimation of the set of hypotheses $\mathcal{H}_{t}^m$ is challenging.  Domain invariant representations have been employed for solving DA problems \cite{MDD_ICML_19,zhao19a,johansson2019support,0080WZLKD021} using Invariant Risk Minimization (IRM) \cite{Arjovsky}. However, recent works elucidated various limitations of IRM. For instance, IRM does not improve accuracy \cite{rosenfeld2021the} or  can lead to worse generalization \cite{KamathTSS21} compared to the standard Empirical Risk Minimization (ERM).

More precisely, Rosenfeld et al. show that unless enough domains covering the space of non-invariant features are observed, a domain invariant nonlinear hypothesis can  underperform on a new test distribution.
Moreover, Kamath et al. show that a sub-optimal hypothesis can be learned using a loss function not being invariant across environments. An attempt to solve this problem is to train models, and selecting domain specific and invariant features by domain specific masks \cite{2020EccvDMG}. However, \citet{gulrajani2021in} show that model selection is non-trivial for domain generalization tasks.

Therefore, we address this problem by composition of three different types of hypotheses:
\begin{itemize}[leftmargin=*]
    \item A \textit{domain dependent} hypothesis function $\bar{h}$ for learning domain specific features.
    \item A \textit{domain invariant} hypothesis function $h^*$ for learning features shared among multiple domains.
    \item A \textit{task-aware} hypothesis function $\hat{h}$ for learning feature representations of tasks, and using these representations to guide information transfer among different domains.
\end{itemize}



Since DNNs can avoid the curse of dimensionality in approximating the class of hierarchically local compositional functions~\cite{Poggio30039}, we approximate hypotheses as follows:

(a) A composite hypothesis ${h = \bar{h} \circ h^* \circ \hat{h}}$ is approximated by parameterized functions $f(\cdot, {\mathcal{W}})$ of a DNN, where ${\mathcal{W}}$ denotes the set of weights of the DNN. 

(b) A component of the hypothesis $h$ is approximated by individual functions $f(\cdot, {w_{b, k}})$, where $w_{b, k} \in \mathcal{W}$ is a weight tensor which will be optimized at the $b^{th}$ branch of the $k^{th}$ segment (i.e. $k^{th}$ group of consecutive layers) of a DNN.


 We use (a) and (b) to define a  compositional framework called Task-guided Multi-branch Zero-shot Domain Adaptation (TG-ZDA) as illustrated in Figure \ref{fig:arc}. We use ERM instead of IRM to estimate hypotheses as suggested in the theoretical results proposed in the recent works. We define and control composability of functions among different domains by multiple branches, and segments (i.e. groups of layers) of the DNN used by the TG-ZDA as depicted in Figure \ref{fig:groupsetting2}.

 \vspace{0.1cm}
 
 \noindent
 \textbf{Identifying Hypotheses by Approximating Functions at Branches and Segments of TG-ZDA } 


TG-ZDA consists of three types of segments, namely (i) domain segment, (ii) shared segment and (iii) task segment. The domain and task segment is domain and task-aware respectively. The shared segment is shared across domains and tasks. The domain segment aims to train hypotheses for using domain specific features and producing domain invariant features. Design of the shared and task segments is inspired by the observations that low-level features are learned at early layers of DNNs, while later layers assemble the learned low-level features into high-level concepts~\cite{lecun2015deeplearning,Sejnowski30033}. The low-level features shared across  tasks are learned by hypotheses at the shared segment. However, characteristics of   low-level features assembled in higher layers  depend on tasks. To capture these characteristics, a task segment is designed for each task. To this end, each $b^{th}$ branch of the $k^{th}$ segment exploits a parameterized approximation $f(\cdot, {w_{b, k} \in \mathcal{W}})$ of a composite hypothesis function $h$ as follows:

\textbf{(i) Domain Segment:} 
While designating a domain segment, we consider that hypotheses $h^d_{sr}$ and $h^d_{t}$ should be estimated successfully for each individual domain using parameterized functions. Hence, functions $f_d^{sr}$ and $f_d^{t}$ are employed at the source and target domain layer using datasets from domains $\mathcal{D}_{sr}$ and $\mathcal{D}_{t}$, respectively. Thereby, \textit{expert} models learned at different branches of TG-ZDA can provide rich representations of individual domains. Therefore, we employ and optimize these functions in different branches of the domain segment without sharing weights.

\textbf{(ii) Shared Segment:} A shared segment is designated, taking inspiration from the observation that low-level features are learned at early layers and the low-level features are universal across tasks~\cite{Saxe11537}. Features obtained from each domain segment $f_d^{\alpha}(x_{\alpha}^{\beta})$, ${\alpha \in \{sr,t\}}$, $\forall \beta \in \{m,a\}$ are fed to the shared segment identifying the function $f_s$. The function $f_s$ is optimized to learn low-level features shared among the main and auxiliary tasks. A shared segment encodes parameters and serves as an inductive bias to improve generalization. 

\textbf{(iii) Task Segment:} As \citet{Bau30071} expound, later layers of DNNs assemble low-level detectors to more sophisticated task-aware semantic representations  ~\cite{Lampinen32970}. Two functions $f_{ta}^m$ and $f_{ta}^a$ are designed to learn representations for main and auxiliary tasks. They are optimized in different branches of the task segment using output features of $f_s$ of the shared segment. 

At last, the features provided by $f_{ta}^m$ and $f_{ta}^a$ are fed to their corresponding classifiers $f_c^{m}$ and $f_c^{a}$, respectively.  

In the training phase, the TG-ZDA takes two datasets of samples $\{ x_{sr}^m \in \mathcal{X}_{sr}^m \}$ and sample pairs ${\{(x_{t}^a,x_{sr}^a) \in \mathcal{X}_{t}^a \times \mathcal{X}_{sr}^a) \}}$ as inputs. Elements in each pair share the same semantic information but from different domains. The TG-ZDA computes class posterior probability of samples belonging to the domain $\alpha \in \{sr,t\}$ for task $\beta \in \{m,a\}$ by
${y_{\alpha}^{\beta}=f_c^{\beta}\circ f_{ta}^{\beta}  \circ f_{s} \circ f_d^{\alpha}(x_{\alpha}^{\beta})}$.

\textbf{An illustrative example of training a TG-ZDA model:} At the right side of Figure \ref{fig:arc}, we give an example of inputs for a task to transfer knowledge of models from a domain of grayscale images~($D_{sr}$) to a domain of colored images~($D_{t}$). Then, the TG-ZDA is used to transfer the knowledge of models on MNIST~\cite{MNIST} with assistance of another task FashionMNIST~\cite{FashionMNIST} (details are given in the next section). Therefore, the main task is defined by classifying samples obtained from MNIST and the auxiliary task is defined by classifying samples belonging to the FashionMNIST and its colored version.

\textbf{Inference using the trained TG-ZDA model:} 
In the inference phase, TG-ZDA takes $x_{t}^m$ as the input and computes its  class posterior probability by
$y_{t}^{m}=f_c^{m}\circ f_{ta}^{m}  \circ f_{s} \circ f_d^{t}(x_{t}^{m})$.

\noindent
\textbf{Loss Functions: } TG-ZDA models are trained using ERM by minimizing the loss function\footnote{Arguments of  functions are not shown to simplify notation.} ${L={\gamma}L_{cls}+L_{D}}$, where $L_{cls}$ is a classification loss and $L_{D}$ is a domain discrepancy loss which are balanced by a hyper-parameter $\gamma>0$. 

\textbf{Classification Loss:} The loss function $L_{cls}$ is defined by 
\begin{equation}
\label{eq:subloss1}
    L_{cls} =\sum_{(\alpha,\beta) \in \{(sr,m),(sr,a)\}}CE(y_{\alpha}^{\beta},\hat{y}_{\alpha}^{\beta}),
\end{equation}
where $y_{\alpha}^{\beta}$ and $\hat{y}_{\alpha}^{\beta}$ is the predicted class posterior probability and its ground truth, respectively for data samples obtained from domain $\alpha$ and task $\beta$, and $CE(\cdot)$  denotes the cross-entropy loss. The loss function \eqref{eq:subloss1} sums classification losses over different tasks and domains. 

\textbf{Domain Discrepancy Loss:} The loss function ${L_{D}= \rho(f_d^{t}(x_{t}^a),f_d^{sr}(x_{sr}^a))}$ is used to measure domain discrepancy $\rho(\cdot, \cdot)$ of representations.
We train TG-ZDA models by minimizing $L_{D}$ for learning domain invariant feature representations. In the next section, we explore TG-ZDA models trained using $\ell_1$ loss, MMD \cite{borgwardt2006integrating} and adversarial loss~\cite{tzeng2017adversarial}. 

\section{Experiments}
\label{sec:exp}

We first give details of experimental setups such as the datasets and models. Next,  TG-ZDA is compared with state-of-the-art methods for multiple ZDA tasks. At last, we examine domain invariance and shareability of features learned using TG-ZDA models. 
Additional results and implementation details are given in the supplemental material. The code will be provided in the camera-ready version of the paper.

\label{sec:dataset}

\begin{table*}
    \begin{center}
    \begin{tabular}{cccc|ccc|cc|cc}
    \toprule
    {Main task} & \multicolumn{3}{c|}{$D_M$} & \multicolumn{3}{c|}{$D_F$} & \multicolumn{2}{c|}{$D_E$} & \multicolumn{2}{c}{$D_N$}\\
    \hline
    {Auxiliary task} & {$D_F$} & {$D_N$} & {$D_E$} & {$D_M$} & {$D_N$} & {$D_E$} & {$D_M$} & {$D_F$} & {$D_M$}  & {$D_F$}\\
    \hline\hline
    \multicolumn{11}{c}{Gray domain to Color domain \textit{(G-dom,C-dom)}.} \\
    \hline\hline
    {ZDDA} & {73.2} & {92.0} & {94.8} & {51.6} & {43.9} & {65.8} & {71.2} & {47.0} & {34.3} & {38.7}\\
    \hline
    {CoCoGAN} & {78.1} & {92.4} & {95.6} & {56.8} & {56.7} & {71.0} & {75.0} & {54.8} & {41.0} & {53.8}\\
    \hline
    {HGNet} & {85.3} & {N/A} & {95.0} & {\color{red} 64.5} & {N/A} & {\color{blue} 71.1} & {71.3} & {57.9} & {N/A} & {N/A}\\
    \hline
    {ALZDA} & {81.2} & {93.3} & {95.0} & {57.4} & {\color{blue} 58.7} & {62.0} & {72.4} & {58.9} & {44.6} & {N/A}\\
    \hline
    {DSPZDA} & {82.9} & {94.4} & {95.3} & {59.1} & {58.6} & {63.1} & {73.6} & {58.3} & {46.4} & {46.5}\\
    \hline
    {LTAZDA} & {81.4} & {94.8} & {95.8} & {62.1} & {\color{red} 60.0} & {\color{red} 74.2} & {79.5} & {63.3} & {46.3} & {49.4}\\
    \hline
    {TG-ZDA-$\ell_1$} & \color{blue}{95.5±3.4} & \color{red}{96.7±0.2} & \color{blue}{99.0±0.0} & \color{blue}{63.4±1.3} & {49.8±1.3} & {65.2±1.6} & \color{red}{90.9±0.5} & \color{red}{87.3±0.3} & \color{red}{75.9±0.4} & \color{red}{61.7±2.7} \\
    \hline
    {TG-ZDA-$mmd$} & \color{red}{97.5±0.2} & \color{blue}{95.9±0.2} & \color{red}{99.1±0.1} & {45.3±0.3} & {32.3±1.7} & {62.2±0.3} & \color{blue}{90.4±0.9} & \color{blue}{82.2±2.9} & \color{blue}{63.0±2.9} & \color{blue}{40.5±4.4} \\
    \hline
    {TG-ZDA-$adv$} & {91.9±0.1} & {95.3±0.0} & {97.8±0.2} & {62.2±4.9} & {49.9±1.2} & {63.3±2.4} & {88.6±0.2} & {71.1±2.4} & {59.2±2.2} & {38.1±4.6} \\
    \hline\hline
    \multicolumn{11}{c}{Gray domain to Edge domain \textit{(G-dom,E-dom)}.} \\
    \hline\hline
    {ZDDA} & {72.5} & {91.5} & {93.2} & {54.1} & {54.0} & {65.8} & {73.6} & {47.0} & {42.3} & {28.4}\\
    \hline
    {CoCoGAN} & {79.6} & {94.9} & {95.4} & {61.5} & {57.5} & {71.0} & {77.9} & {54.8} & {48.0} & {36.3}\\
    \hline
    {HGNet} & {86.5} & {N/A} & {96.1} & {N/A} & {N/A} & {N/A} & {81.1} & {57.9} & {N/A} & {N/A}\\
    \hline
    {ALZDA} & {81.4} & {93.5} & {96.3} & {63.2} & {58.7} & {72.4} & {78.2} & {58.9} & {49.9} & {N/A}\\
    \hline
    {DSPZDA} & {84.7} & {93.9} & {95.5} & \color{blue}{64.8} & \color{blue}{60.0} & \color{blue}{73.3} & {79.2} & {64.7} & {50.2} & {43.3}\\
    \hline
    {LTAZDA} & {81.1} & {95.0} & {95.6} & \color{red}{66.3} & \color{red}{60.7} & \color{red}{74.6} & {78.4} & {60.6} & {54.8} & \color{red}{51.3}\\
    \hline
    {TG-ZDA-$\ell_1$} & \color{red}{97.0±0.2} & \color{red}{96.1±0.5} & \color{red}{99.2±0.0} & {62.8±2.6} & {47.5±1.3} & {66.2±2.3} & \color{red}{91.0±0.4} & \color{blue}{73.2±4.1} & \color{red}{62.3±1.3} & \color{blue}{49.0±1.7} \\
    \hline
    {TG-ZDA-$mmd$} & \color{blue}{96.6±0.6} & {95.2±0.4} & \color{blue}{99.1±0.0} & {53.1±4.2} & {31.4±2.4} & {67.6±2.3} & \color{blue}{89.3±0.3} & \color{red}{75.1±2.3} & \color{blue}{58.7±1.8} & {32.9±1.2} \\
    \hline
    {TG-ZDA-$adv$} & {93.8±1.7} & \color{blue}{95.9±0.4} & {98.0±0.3} & {54.6±6.3} & {48.9±3.3} & {59.2±1.0} & {87.5±0.3} & {70.9±3.3} & {49.0±2.5} & {19.4±3.8} \\

    \hline\hline
    \multicolumn{11}{c}{Gray domain to Negative domain \textit{(G-dom,N-dom)}.} \\
    \hline\hline
    {ZDDA} & {77.9} & {82.4} & {90.5} & {61.4} & {47.4} & {65.3} & {76.2} & {53.4} & {37.8} & {38.7}\\
     \hline
    {CoCoGAN} & {80.3} & {87.5} & {93.1} & {66.0} & {52.2} & {66.8} & {81.1} & {56.5} & {45.7} & {53.8}\\
     \hline
    {HGNet} & {83.7} & {N/A} & {95.7} & {N/A} & {N/A} & {71.1} & {N/A} & {62.3} & {N/A} & {N/A}\\
     \hline
    {ALZDA} & {N/A} & {N/A} & {N/A} & {N/A} & {N/A} & {N/A} & {N/A} & {N/A} & {N/A} & {N/A}\\
    \hline
    {DSPZDA} & {83.6} & {89.1} & {94.2} & {71.1} & {58.0} & {70.3} & {82.1} & {65.2} & {49.6} & {57.4}\\
    \hline
    {LTAZDA} & {81.6} & {90.4} & {94.2} & {68.3} & {61.5} & {70.8} & {81.5} & {60.3} & {52.9} & {56.4}\\
    \hline
    {TG-ZDA-$\ell_1$} & \color{blue}{99.2±0.1} & \color{red}{99.4±0.0} & \color{red}{99.4±0.0} & \color{red}{82.4±0.5} & \color{blue}{83.0±1.3} & \color{red}{83.6±0.2} & \color{blue}{92.3±0.2} & \color{blue}{91.7±0.2} & \color{blue}{84.9±0.4} & \color{blue}{85.0±0.2} \\
    \hline
    {TG-ZDA-$mmd$} & \color{red}{99.3±0.0} & \color{blue}{99.4±0.0} & {99.3±0.1} & \color{blue}{82.1±0.9} & \color{red}{83.5±0.7} & \color{blue}{83.2±0.3} & \color{red}{92.7±0.1} & \color{red}{91.9±0.4} & \color{red}{85.6±0.2} & \color{red}{85.3±0.2} \\
    \hline
    {TG-ZDA-$adv$} & {99.1±0.2} & {99.2±0.1} & \color{blue}{99.4±0.0} & {80.4±1.0} & {77.9±2.7} & {80.0±0.4} & {91.6±0.1} & {90.4±0.2} & {73.4±2.1} & {65.8±2.5} \\

    \hline\hline
    \multicolumn{11}{c}{Color domain to Gray domain \textit{(C-dom,G-dom)}.} \\
    \hline\hline
    {ZDDA} & {67.4} & {85.7} & {87.6} & {55.1} & {49.2} & {59.5} & {39.6} & {23.7} & {75.5} & {52.0}\\
     \hline
    {CoCoGAN} & {73.2} & {89.6} & {94.7} & {61.1} & {50.7} & {70.2} & {47.5} & {57.7} & {80.2} & {67.4}\\
    \hline
    {HGNet} & {78.9} & {N/A} & {95.0} & {65.9} & {N/A} & {68.5} & {N/A} & {N/A} & {N/A} & {N/A}\\
    \hline
    {ALZDA} & {73.4} & {91.0} & {93.4} & {62.4} & {53.5} & {71.5} & {50.6} & {58.1} & {83.5} & {70.9}\\
    \hline
    {DSPZDA} & {76.2} & {94.6} & {95.2} & {63.2} & {54.3} & {73.6} & {84.3} & {71.5} & {53.8} & {61.4}\\
    \hline
    {LTAZDA} & {73.8} & {93.1} & {95.0} & {66.1} & {54.1} & {74.3} & {84.0} & {77.2} & {56.4} & {63.6}\\
    \hline
    {TG-ZDA-$\ell_1$} & \color{red}{98.7±0.2} & \color{red}{98.7±0.1} & \color{red}{99.0±0.0} & \color{red}{74.2±1.7} & \color{red}{74.8±1.2} & \color{red}{77.4±2.1} & \color{red}{85.6±0.7} & \color{red}{80.3±0.4} & \color{red}{84.0±0.2} & \color{red}{82.7±1.0} \\
    \hline
    {TG-ZDA-$mmd$} & \color{blue}{97.8±0.1} & \color{blue}{98.4±0.1} & \color{blue}{98.4±0.2} & \color{blue}{74.0±3.4} & \color{blue}{72.1±2.8} & {70.2±4.8} & \color{blue}{81.9±1.1} & {71.0±1.1} & \color{blue}{83.5±0.5} & \color{blue}{81.5±0.2} \\
    \hline
    {TG-ZDA-$adv$} & {96.4±1.3} & {96.8±0.9} & {98.3±0.1} & {70.6±2.0} & {58.2±6.1} & \color{blue}{72.5±2.1} & {80.6±0.8} & \color{blue}{77.7±2.9} & {68.9±2.5} & {67.7±3.9} \\

    \hline\hline
    \multicolumn{11}{c}{Negative domain to Gray domain \textit{(N-dom,G-dom)}.} \\
    \hline\hline
    {ZDDA} & {78.5} & {90.7} & {87.6} & {56.6} & {57.1} & {67.1} & {34.1} & {39.5} & {67.7} & {45.5}\\
     \hline
    {CoCoGAN} & {80.1} & {92.8} & {93.6} & {63.4} & {61.0} & {72.8} & {47.0} & {43.9} & {78.8} & {58.4}\\
    \hline
    {HGNet} & {87.5} & {N/A} & {95.0} & {64.6} & {N/A} & {75.1} & {78.0} & {67.9} & {N/A} & {N/A}\\
    \hline
    {ALZDA} & {82.6} & {94.6} & {95.8} & {67.0} & {68.2} & \color{red}{77.9} & {51.1} & {44.2} & {79.7} & {62.2}\\
    \hline
    {TG-ZDA-$\ell_1$} & \color{blue}{98.7±0.4} & \color{blue}{98.3±0.4} & \color{red}{99.1±0.0} & \color{red}{76.5±0.7} & \color{blue}{74.7±1.1} & {74.2±1.0} & \color{red}{85.6±0.5} & \color{blue}{84.0±0.6} & \color{blue}{85.0±0.5} & \color{red}{85.6±0.2} \\
    \hline
    {TG-ZDA-$mmd$} & \color{red}{99.0±0.0} & \color{red}{98.8±0.0} & \color{blue}{99.0±0.1} & \color{blue}{76.3±1.2} & \color{red}{75.7±0.4} & {75.2±1.6} & {84.9±1.4} & \color{red}{84.2±0.3} & \color{red}{85.7±0.2} & \color{blue}{85.4±0.2} \\
    \hline
    {TG-ZDA-$adv$} & {98.6±0.0} & {97.3±0.3} & {99.0±0.1} & {76.0±0.8} & {60.0±3.0} & \color{blue}{75.5±1.5} & \color{blue}{85.4±1.8} & {82.8±0.9} & {73.1±1.7} & {69.3±2.2} \\
    \bottomrule
    \end{tabular}
    \end{center}
     \caption{Comparison of accuracy (\%) of methods. N/A indicates that accuracy is not reported by the corresponding method ({\color{red} \textbf{red}} and {\color{blue} \textbf{blue}} indicates the best and the second best accuracy.). Since the accuracy values of DSPZDA and LTAZDA are not reported in the corresponding papers for \textit{(N-dom,G-dom)}, they are not shown in the table.}
     \label{tab:n2g}
\end{table*}


\textbf{Datasets: } Our proposed TG-ZDA is examined on four datasets; MNIST \cite{MNIST}, FashionMNIST \cite{FashionMNIST}, NIST \cite{NIST} and EMNIST \cite{EMNIST}  and their variants, following the experimental settings proposed in \cite{Peng2018ZeroShotDD,Wang2019ConditionalCG}. The four datasets are denoted by $D_M$, $D_F$, $D_N$ and $D_E$ respectively. Their original versions are in gray domain \textit{(G-dom)}.  The color domain \textit{(C-dom)}, the edge domain \textit{(E-dom)} and the negative domain \textit{(N-dom)}, are created for each dataset for evaluation.


\begin{figure*}
\centering
\subfigure[$x_t^{m} \& x_{sr}^{m},m=D_M$.]{
\includegraphics[width=0.18\textwidth]{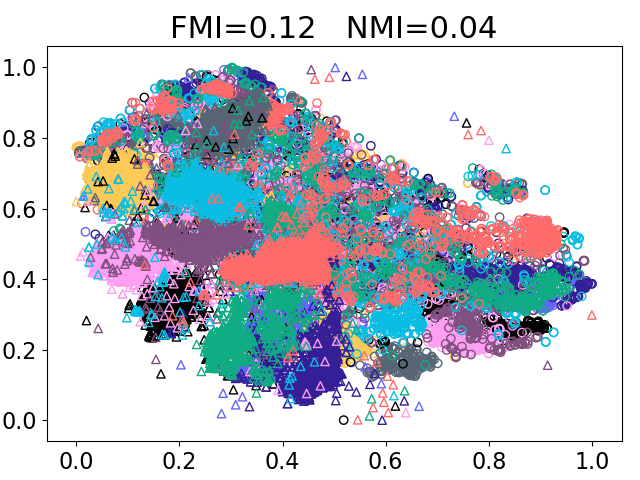}
\label{subfig:tsne_img_dm}}
\subfigure[$q(x),a=D_E$.]{
\includegraphics[width=0.18\textwidth]{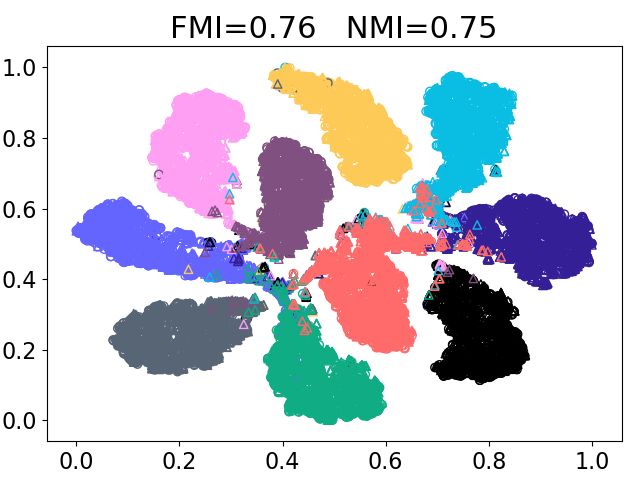}
\label{subfig:tsne_fcin_dm_de}}
\subfigure[$q(x),a=D_F$.]{
\includegraphics[width=0.18\textwidth]{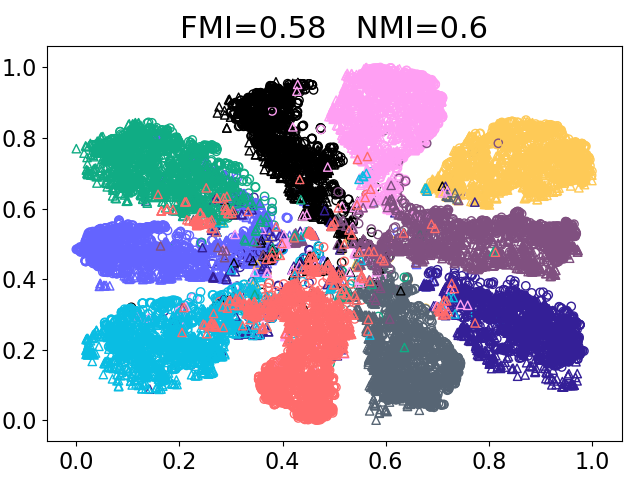}
\label{subfig:tsne_fcin_dm_df}}
\subfigure[$q(x),a=D_N$.]{
\includegraphics[width=0.18\textwidth]{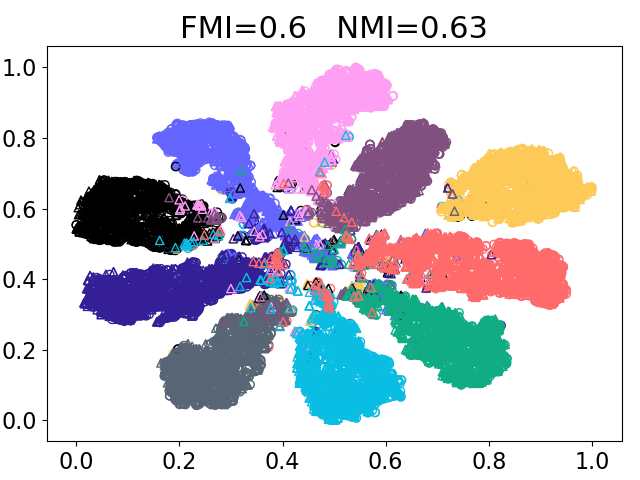}
\label{subfig:tsne_fcin_dm_dn}}
\subfigtopskip=-5pt

\subfigure[$x_t^{m} \& x_{sr}^{m},m=D_F$.]{
\includegraphics[width=0.18\textwidth]{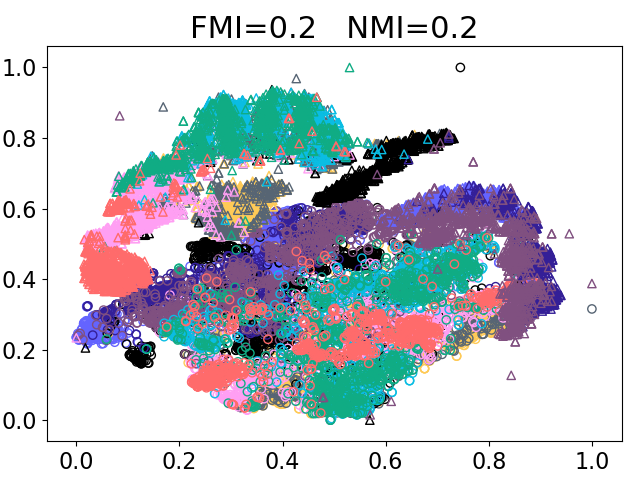}
\label{subfig:tsne_img_df}}
\subfigure[$q(x),a=D_E$.]{
\includegraphics[width=0.18\textwidth]{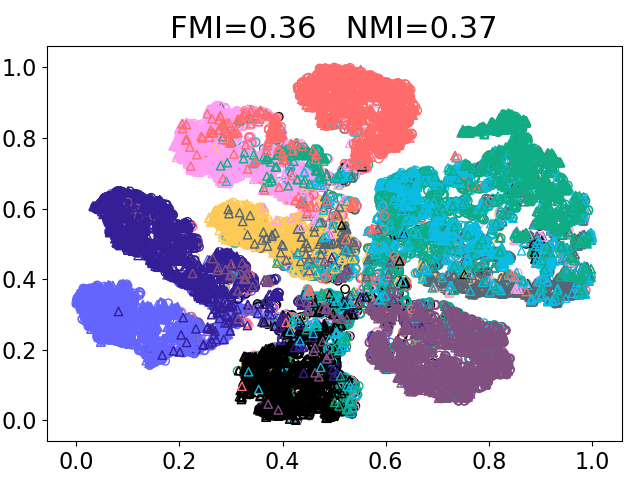}
\label{subfig:tsne_fcin_df_de}}
\subfigure[$q(x),a=D_M$.]{
\includegraphics[width=0.18\textwidth]{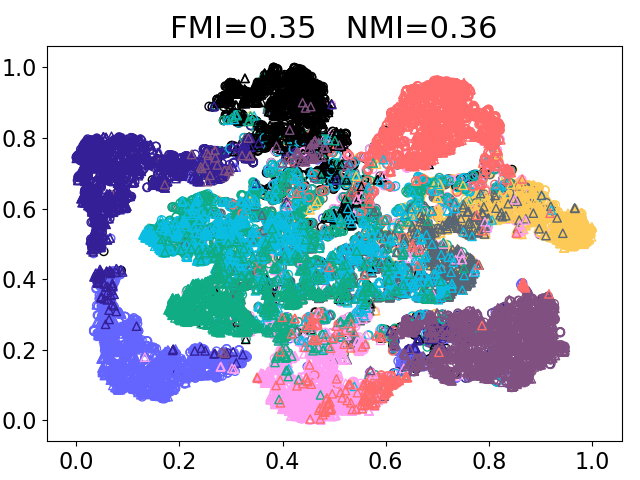}
\label{subfig:tsne_fcin_df_dm}}
\subfigure[$q(x),a=D_N$.]{
\includegraphics[width=0.18\textwidth]{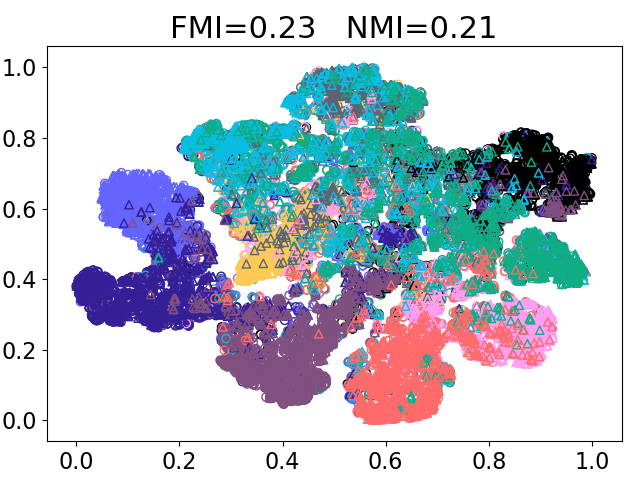}
\label{subfig:tsne_fcin_df_dn}}
\subfigure[Class labels.]{
\includegraphics[width=0.18\textwidth,height=0.15\textwidth]{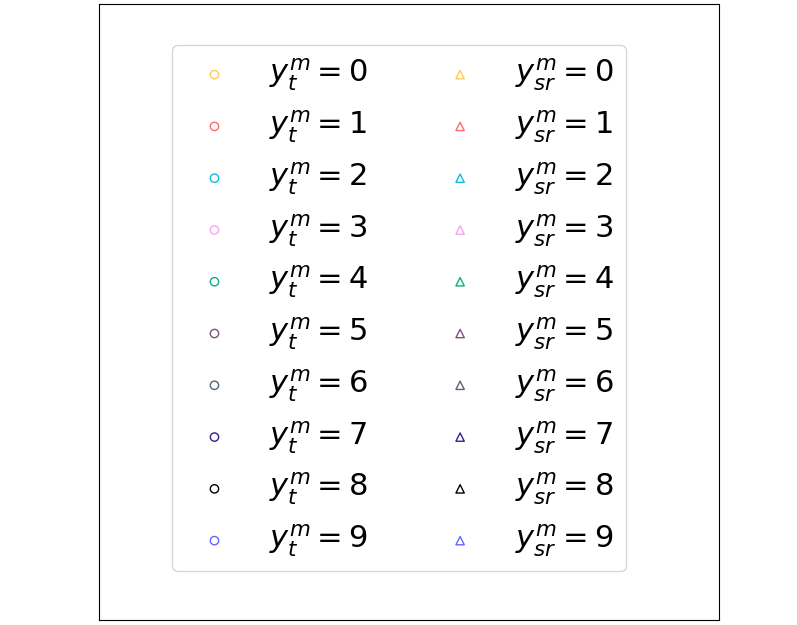}
\label{subfig:legend2}}
\vspace{-0.3cm}
\caption{Visualization of distribution of representations using t-SNE when the domain pair is \textit{(G-dom,C-dom)}. The figures in the first row show features learned when the main task is $D_M$ while the auxiliary task is (b) $D_E$, (c) $D_F$ and (d) $D_N$. Similarly, the second row shows figures when the main task is $D_F$ while the auxiliary task is (f) $D_E$, (g) $D_M$ and (h) $D_N$. Figures (a) and (e) show distributions of input data $x_t^{m}$. Classes of $x_t^{m}$ are indicated by different colors. Figures (b)-(d) and (f)-(h) show distributions of features computed by $q(x)$ for the corresponding auxiliary tasks, where $q \triangleq f_{
ta}^{m} \circ f_{s} \circ f_d^{\alpha},\alpha \in \{sr, t\}$, and $x$ denotes either $x_t^{m}$ or $x_{sr}^{m}$. Triangle and circle markers indicate features belonging to the source and target domains. }
\label{fig:tsne}
\end{figure*}

\textbf{Network Architecture:} 
We use a VGG16 \cite{VGG16} with instance normalization (IN) layers denoted by VGG16-IN as the backbone network, and use its different partitions  as the segments of the TG-ZDA network. 
\citet{Huang2017ArbitraryST} showed that the mean and variance of features encode style information of input images.  Since style information is highly domain dependent, we consider that IN can be used to extract domain invariant features by statistically normalizing features, helping to disentangle category specific (semantic) and domain specific features. 

\textbf{Design of segments of TG-ZDA:} The VGG16-IN consists of five blocks, which are split into three groups by two hyperparameters $s_1$ and $s_2$. The setting ${(s_1,s_2)=(1,2)}$ indicates that the VGG-IN is separated into three segments at the first and second maxpooling layer. Each group is used as a segment of the TG-ZDA network. The second and third row of Figure \ref{fig:groupsetting2}-(a) shows an illustration of the three segments obtained for $(s_1,s_2)=(1,2)$ and $(s_1,s_2)=(2,3)$. The ReLu is omitted in the figure. The bottom row of Figure~\ref{fig:groupsetting2} shows the architecture for $(s_1,s_2)=(2,3)$, which is the default setting used in the  experiments. 
 All classifiers $f_c^{\beta},\beta \in\{m,a\}$ consist of three fully connected layers. 

\noindent
\textbf{Comparison with State-of-the-art}

Akin to \cite{Wang2019ConditionalCG}, we evaluate TG-ZDA for five pairs of source and target domains. \textit{G-dom} is set as the source domain in three of them. They are \textit{(G-dom, C-dom)}, \textit{(G-dom, E-dom)}, and \textit{(G-dom, N-dom)}. The other two pairs take \textit{G-dom} as the target domain, where the domain pairs are \textit{(C-dom, G-dom)} and \textit{(N-dom, G-dom)}. For each domain pair, we evaluate TG-ZDA on four different tasks corresponding to the four datasets given in the previous section. Each task consists of a main and an auxiliary task. For instance, when a given task is domain adaptation from \textit{G-dom} to \textit{C-dom} on the MNIST dataset, the FashionMNIST can be used to assist TG-ZDA to perform the task. In this case, the main task classifies  MNIST digits (\textit{G-dom}), while the auxiliary task classifies FashionMNIST clothes belonging to \textit{G-dom} and \textit{C-dom}. \textit{($D_N$, $D_E$)} and \textit{($D_E$, $D_N$)} are not viewed as valid (\textit{main task}, \textit{auxiliary task}) pairs in our experiments, since they both contain images of letters. We compare TG-ZDA, optimized by $\ell_1$ loss, MMD  loss~\cite{borgwardt2006integrating} and adversarial loss~\cite{tzeng2017adversarial} respectively, with six state-of-the-art methods, ZDDA \cite{Peng2018ZeroShotDD, zddaTpami}, CoCoGAN \cite{Wang2019ConditionalCG}, ALZDA \cite{wang2020adversarial}, LATZDA \cite{LATZDA}, HGNet \cite{Xia2020HGNetHG} and DSPZDA \cite{DSPZDA}. TG-ZDA models trained by $\ell_1$, MMD and adversarial loss is denoted by TG-ZDA-$\ell_1$, TG-ZDA-$mmd$ and TG-ZDA-$adv$.

Table \ref{tab:n2g} provides mean (±standard deviation) classification accuracy over three random trials. 
In the analyses, TG-ZDA achieved top two accuracy in 44 tasks and top one accuracy in 42 tasks out of 50 tasks. 
Albation study of loss functions shows that TG-ZDA-$\ell_1$ obtained the top-two-accuracy in 44 tasks, and TG-ZDA-$mmd$ achieved top-two-accuracy in 37 tasks. 
We conjecture that the superiority of TG-ZDA-$\ell_1$ and TG-ZDA-$mmd$ over TG-ZDA-$adv$ is due primarily to the fact that TG-ZDA-$\ell_1$ and TG-ZDA-$mmd$ can learn intermediate domain invariant feature representation by estimating hypotheses at both the source and target branch simultaneously, while TG-ZDA-$adv$ matches  distribution of features learned at the target domain branch to that of the source domain branch. Even though TG-ZDA-$adv$ performs the worst among three TG-ZDA variants, it still outperforms the state-of-the-art methods by a large margin in 34 tasks, demonstrating effectiveness of TG-ZDA.


\noindent
\textbf{Analyses of Learned Feature Representations}

\begin{figure}
\centering
\subfigure[Domain invariant features.]{
\includegraphics[width=0.48\textwidth]{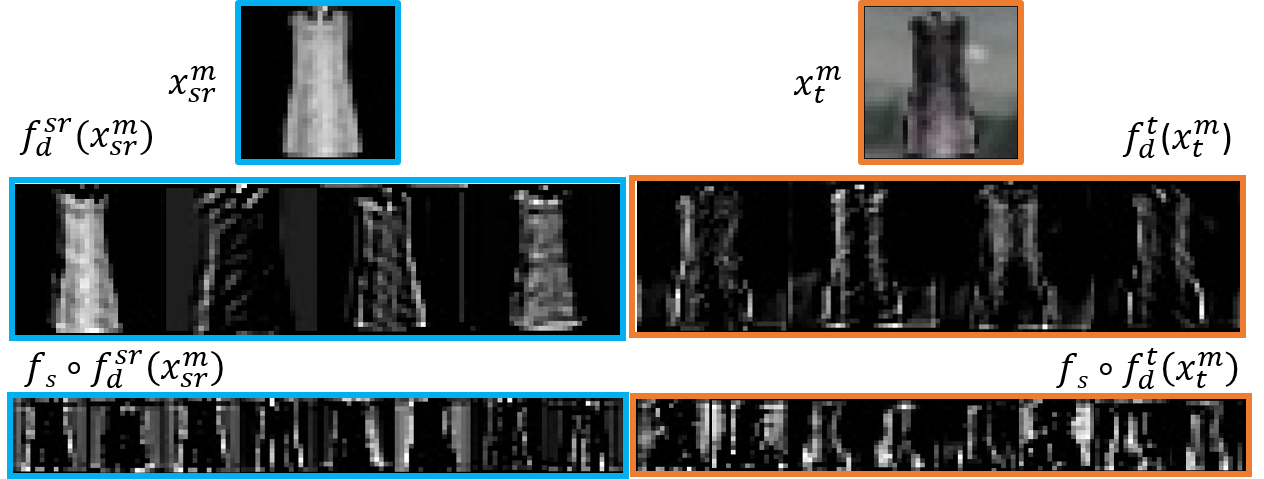}
\label{fig:featurevis}}
\subfigure[Shareability of features among domains.]{
\includegraphics[width=0.48\textwidth,height=0.15\textwidth]{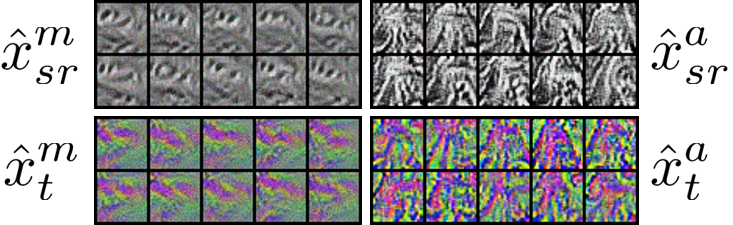}
\label{subfig:x_opt}}
\vspace{-0.3cm}
\caption{(a) Domain invariant features learned by TG-ZDA when the main task is $D_F$, assisted by $D_M$. (b) Visualization of features $\hat{x}_{sr}^{m}$,  $\hat{x}_{t}^{m}$, $\hat{x}_{sr}^{a}$ and $\hat{x}_{t}^{a}$ shared among domains.}
\end{figure}

\textbf{Feature separability:} We visualize representations learned using TG-ZDA with t-SNE \cite{tsne} and elucidate their effectiveness. Figure~\ref{fig:tsne} shows results for the domain pair \textit{(G-dom, C-dom)} as the main task is $D_{M}$ and $D_F$. Figure \ref{subfig:tsne_img_dm} and \ref{subfig:tsne_img_df} depict distributions of $x_t^{m}$ from unseen target domain, where its main task is $D_{M}$ and $D_F$, respectively. The results show that the samples of $x_t^{m}$ are not well separable among classes. In contrast, class separability improves in the feature space provided by $q \triangleq f_{ta}^{m}  \circ f_{s} \circ f_d^{\alpha},\alpha \in \{sr, t\}$. They are indicated by circle markers in Figure \ref{subfig:tsne_fcin_dm_de}-\ref{subfig:tsne_fcin_dm_dn} and \ref{subfig:tsne_fcin_df_de}-\ref{subfig:tsne_fcin_df_dn} which demonstrate that the segments of TG-ZDA improve separability of $x_t^{m}$. However, the results do not indicate that the features of $x_t^{m}$ can be classified well by $f_c^{m}$ since $f_c^{m}$ is trained using $x_{sr}^{m}$. If features of $x_{sr}^{m}$ and $x_t^{m}$ are \textit{well} discriminated, then the features of $x_t^{m}$ can be classified \textit{well} by $f_c^{m}$ trained using $x_{sr}^{m}$. Triangle markers in Figure \ref{subfig:tsne_fcin_dm_de}-\ref{subfig:tsne_fcin_dm_dn} and \ref{subfig:tsne_fcin_df_de}-\ref{subfig:tsne_fcin_df_dn} demonstrate this property. Triangle and circle markers indicate representations of $x_{sr}^{m}$ and $x_t^{m}$, respectively. Colors of the markers differ by class id. We observe that representations of $x_{sr}^{m}$ and $x_t^{m}$ belonging to the same class are well-clustered, demonstrating \textit{good} feature separability among classes. 

\textbf{Alignment of features among domains:} We analyze alignment of learned features by first clustering features belonging to the source and target domain using k-means \cite{kmeans}. Then, we compute Fowlkes-Mallows index (FMI) \cite{FMI} and normalized mutual information (NMI) \cite{muller2016introduction} for clusters  belonging to the source and target domain. 
FMI $\in [0,1]$ measures alignment of features between target and source domain considering their class labels, and higher FMI values indicate better alignment. NMI $\in [0,1]$  measures statistical correlation among clusters of features obtained from different domains, and higher values indicate better correlation. 

In Figure~\ref{fig:tsne}, features obtained using composite hypotheses provide higher FMI and NMI values. This result shows  that feature alignment among different domains is improved by  task guided training of compositional hypotheses of representations. More precisely, FMI and NMI computed using $x_t^{m}$ and $x_{sr}^m$ is $0.12$ and $0.04$ respectively, revealing a bad alignment in Figure~\ref{subfig:tsne_img_dm} and~\ref{subfig:tsne_img_df}. They increase to $0.76$ and $0.75$, respectively, indicating a \textit{good} alignment, after mapping $x_t^{m}$ and $x_{sr}^m$ to feature spaces by $q(x)$. Moreover, the results show that a better cluster and feature alignment results in better classification by comparing Figure~\ref{subfig:tsne_fcin_dm_de} and \ref{subfig:tsne_fcin_df_de}, and their corresponding accuracy in Table~\ref{tab:n2g}.

\textbf{Domain invariance and feature shareability:}
We visualize feature maps learned using TG-ZDA by optimising $x^{\beta}_{\alpha}$ to examine the learned domain invariants and benefits of feature shareability among domains. In the analyses, we used \textit{(G-dom, C-dom)} as the \textit{(source, target)} pair. We first depict  outputs of the domain and shared segments in Figure \ref{fig:featurevis}, when the main and auxiliary task is classifying samples from FashionMNIST and MNIST, respectively. The feature maps depict the learned task-aware semantic information. Next, we explore the benefits of learning features shared among tasks and domains using TG-ZDA  by visualizing $x^{\beta}_{\alpha}$ which maximize the output $y^{\beta}_{\alpha},\alpha \in \{sr,t\},\beta \in \{m,a\}$. 
Through optimization, it searches for inputs $\hat{x}_{\alpha}^{\beta}$ that make neurons of the last layer fire with largest values. To explore sharability properties, we visualize $\hat{x}^{\beta=m}_{\alpha=t}$ and $\hat{x}^{\beta=a}_{\alpha=sr}$, since neither of their corresponding hypotheses have access to ${x}_{\alpha}^{\beta}$. We observe that $\hat{x}^{\beta}_{\alpha}$ is in the domain $\alpha$ and contains patterns from the task $\beta$. This result suggests that ZDA is established through sharability of features among domains and tasks.


In addition, we provide the results when the main task is classifying digits (MNIST) and the auxiliary task is classifying clothes (FashionMNIST), where $sr$ is gray domain, $t$ is color domain, $m$ is classifying MNIST and $a$ is classifying FashionMNIST. Figure \ref{subfig:x_opt} shows $\hat{x}_{sr}^{m}$, $\hat{x}_{t}^{m}$, $\hat{x}_{sr}^{a}$ and $\hat{x}_{t}^{a}$. They are depicted with feature representations of patterns learned in their corresponding domains. For instance, $\hat{x}_{sr}^{m}$ shows learned patterns of digits in the grayscale domain (MNIST), and $\hat{x}_{t}^{a}$ shows learned patterns of clothes in the color domain (FashionMNIST). These results are admissible and expected due to their optimizing objectives and access of corresponding data. However, $\hat{x}_t^{m}$, shown in Figure \ref{subfig:x_opt}, contains  patterns of digits in the color domain, whose data are not accessed during training. These results show that TG-ZDA can learn representations of MNIST digits in color domain, in the manner of zero-shot, although it is never trained by images of digits from the target color domain.

\section{Conclusion}
ZDA problem is defined as training models where the target test data cannot be accessed in training phase. We have proposed a novel framework called TG-ZDA which employs a multi-branch DNN to solve this problem. The proposed TG-ZDA disentangles ZDA hypotheses to three functions. Each function is realized by a branch of a segment in TG-ZDA. Branches in the segments are recombined according to tasks. TG-ZDA learns domain specific features and sharealibility to overcome the absence of target data. Experimental results demonstrate that our proposed TG-ZDA outperforms state-of-the-art methods in benchmark ZDA tasks, validating the effectiveness of TG-ZDA.

\bibliography{egbib}

\begin{thebibliography}{53}
\providecommand{\natexlab}[1]{#1}

\bibitem[{Arjovsky et~al.(2020)Arjovsky, Bottou, Gulrajani, and
  Lopez-Paz}]{Arjovsky}
Arjovsky, M.; Bottou, L.; Gulrajani, I.; and Lopez-Paz, D. 2020.
\newblock Invariant Risk Minimization.
\newblock arXiv:1907.02893.

\bibitem[{Bau et~al.(2020)Bau, Zhu, Strobelt, Lapedriza, Zhou, and
  Torralba}]{Bau30071}
Bau, D.; Zhu, J.-Y.; Strobelt, H.; Lapedriza, A.; Zhou, B.; and Torralba, A.
  2020.
\newblock Understanding the role of individual units in a deep neural network.
\newblock \emph{Proceedings of the National Academy of Sciences}, 117(48):
  30071--30078.

\bibitem[{Borgwardt et~al.(2006)Borgwardt, Gretton, Rasch, Kriegel,
  Sch{\"o}lkopf, and Smola}]{borgwardt2006integrating}
Borgwardt, K.~M.; Gretton, A.; Rasch, M.~J.; Kriegel, H.-P.; Sch{\"o}lkopf, B.;
  and Smola, A.~J. 2006.
\newblock Integrating structured biological data by kernel maximum mean
  discrepancy.
\newblock \emph{Bioinformatics}, 22(14): e49--e57.

\bibitem[{Bousmalis et~al.(2017)Bousmalis, Silberman, Dohan, Erhan, and
  Krishnan}]{bousmalis2017unsupervised}
Bousmalis, K.; Silberman, N.; Dohan, D.; Erhan, D.; and Krishnan, D. 2017.
\newblock Unsupervised pixel-level domain adaptation with generative
  adversarial networks.
\newblock In \emph{Proceedings of the IEEE Conference on Computer Vision and
  Pattern Recognition (CVPR)}, 3722--3731.

\bibitem[{Chattopadhyay, Balaji, and Hoffman(2020)}]{2020EccvDMG}
Chattopadhyay, P.; Balaji, Y.; and Hoffman, J. 2020.
\newblock Learning to Balance Specificity and Invariance for In and Out of
  Domain Generalization.
\newblock In Vedaldi, A.; Bischof, H.; Brox, T.; and Frahm, J.-M., eds.,
  \emph{Proceedings of the European Conference on Computer Vision (ECCV)},
  301--318.

\bibitem[{Chen et~al.(2020)Chen, Fu, Chen, Jin, Cheng, Jin, and
  Hua}]{Chen2020HoMMHM}
Chen, C.; Fu, Z.; Chen, Z.; Jin, S.; Cheng, Z.; Jin, X.; and Hua, X. 2020.
\newblock HoMM: Higher-order Moment Matching for Unsupervised Domain
  Adaptation.
\newblock In \emph{Proceedings of the AAAI Conference on Artificial
  Intelligence}, 3422--3429.

\bibitem[{Cohen et~al.(2017)Cohen, Afshar, Tapson, and van Schaik}]{EMNIST}
Cohen, G.; Afshar, S.; Tapson, J.; and van Schaik, A. 2017.
\newblock EMNIST: Extending MNIST to handwritten letters.
\newblock In \emph{2017 International Joint Conference on Neural Networks
  (IJCNN)}, 2921--2926.

\bibitem[{Creager, Jacobsen, and Zemel(2021)}]{creager21environment}
Creager, E.; Jacobsen, J.-H.; and Zemel, R. 2021.
\newblock Environment Inference for Invariant Learning.
\newblock In \emph{International Conference on Machine Learning}.

\bibitem[{Fowlkes and Mallows(1983)}]{FMI}
Fowlkes, E.~B.; and Mallows, C.~L. 1983.
\newblock A Method for Comparing Two Hierarchical Clusterings.
\newblock \emph{Journal of the American Statistical Association}, 78(383):
  553--569.

\bibitem[{Ganin et~al.(2016)Ganin, Ustinova, Ajakan, Germain, Larochelle,
  Laviolette, Marchand, and Lempitsky}]{ganin2016domain}
Ganin, Y.; Ustinova, E.; Ajakan, H.; Germain, P.; Larochelle, H.; Laviolette,
  F.; Marchand, M.; and Lempitsky, V. 2016.
\newblock Domain-adversarial training of neural networks.
\newblock \emph{The Journal of Machine Learning Research}, 17(1): 2096--2030.

\bibitem[{Grother(1970)}]{NIST}
Grother, P. 1970.
\newblock NIST Special Database 19. NIST Handprinted Forms and Characters
  Database.

\bibitem[{Gulrajani and Lopez-Paz(2020)}]{gulrajani2021in}
Gulrajani, I.; and Lopez-Paz, D. 2020.
\newblock In Search of Lost Domain Generalization.
\newblock arXiv:2007.01434.

\bibitem[{Huang et~al.(2018)Huang, Lin, Chen, Wu, Hsu, and
  Lai}]{huang2018auggan}
Huang, S.-W.; Lin, C.-T.; Chen, S.-P.; Wu, Y.-Y.; Hsu, P.-H.; and Lai, S.-H.
  2018.
\newblock Auggan: Cross domain adaptation with gan-based data augmentation.
\newblock In \emph{Proceedings of the European Conference on Computer Vision
  (ECCV)}, 718--731.

\bibitem[{Huang and Belongie(2017)}]{Huang2017ArbitraryST}
Huang, X.; and Belongie, S.~J. 2017.
\newblock Arbitrary Style Transfer in Real-Time with Adaptive Instance
  Normalization.
\newblock \emph{Proceedings of the IEEE International Conference on Computer
  Vision (ICCV)}, 1510--1519.

\bibitem[{Johansson, Sontag, and Ranganath(2019)}]{johansson2019support}
Johansson, F.~D.; Sontag, D.; and Ranganath, R. 2019.
\newblock Support and invertibility in domain-invariant representations.
\newblock In \emph{The 22nd International Conference on Artificial Intelligence
  and Statistics}, 527--536.

\bibitem[{Kamath et~al.(2021)Kamath, Tangella, Sutherland, and
  Srebro}]{KamathTSS21}
Kamath, P.; Tangella, A.; Sutherland, D.; and Srebro, N. 2021.
\newblock Does Invariant Risk Minimization Capture Invariance?
\newblock In \emph{Proceedings of The 24th International Conference on
  Artificial Intelligence and Statistics}, 4069--4077.

\bibitem[{Kutbi, Peng, and Wu(2021)}]{zddaTpami}
Kutbi, M.; Peng, K.-C.; and Wu, Z. 2021.
\newblock Zero-shot Deep Domain Adaptation with Common Representation Learning.
\newblock \emph{IEEE Transactions on Pattern Analysis and Machine
  Intelligence}, 1--1.

\bibitem[{Lampinen and McClelland(2020)}]{Lampinen32970}
Lampinen, A.~K.; and McClelland, J.~L. 2020.
\newblock Transforming task representations to perform novel tasks.
\newblock \emph{Proceedings of the National Academy of Sciences}, 117(52):
  32970--32981.

\bibitem[{LeCun, Bengio, and Hinton(2015)}]{lecun2015deeplearning}
LeCun, Y.; Bengio, Y.; and Hinton, G. 2015.
\newblock Deep Learning.
\newblock \emph{Nature}, 521(7553): 436--444.

\bibitem[{Lecun et~al.(1998)Lecun, Bottou, Bengio, and Haffner}]{MNIST}
Lecun, Y.; Bottou, L.; Bengio, Y.; and Haffner, P. 1998.
\newblock Gradient-based learning applied to document recognition.
\newblock \emph{Proceedings of the IEEE}, 86(11): 2278--2324.

\bibitem[{Lee et~al.(2019)Lee, Batra, Baig, and Ulbricht}]{Lee2019SlicedWD}
Lee, C.-Y.; Batra, T.; Baig, M.~H.; and Ulbricht, D. 2019.
\newblock Sliced Wasserstein Discrepancy for Unsupervised Domain Adaptation.
\newblock In \emph{Proceedings of the IEEE Conference on Computer Vision and
  Pattern Recognition (CVPR)}, 10277--10287.

\bibitem[{Li et~al.(2021)Li, Wang, Zhang, Li, Keutzer, Darrell, and
  Zhao}]{0080WZLKD021}
Li, B.; Wang, Y.; Zhang, S.; Li, D.; Keutzer, K.; Darrell, T.; and Zhao, H.
  2021.
\newblock Learning Invariant Representations and Risks for Semi-Supervised
  Domain Adaptation.
\newblock In \emph{Proceedings of the IEEE Conference on Computer Vision and
  Pattern Recognition (CVPR)}, 1104--1113.

\bibitem[{Liang, Hu, and Feng(2020)}]{liang2020shot}
Liang, J.; Hu, D.; and Feng, J. 2020.
\newblock Do We Really Need to Access the Source Data? Source Hypothesis
  Transfer for Unsupervised Domain Adaptation.
\newblock In \emph{International Conference on Machine Learning (ICML)},
  6028--6039.

\bibitem[{Liu et~al.(2019{\natexlab{a}})Liu, Long, Wang, and
  Jordan}]{Liu2019TransferableAT}
Liu, H.; Long, M.; Wang, J.; and Jordan, M. 2019{\natexlab{a}}.
\newblock Transferable Adversarial Training: A General Approach to Adapting
  Deep Classifiers.
\newblock In \emph{Proceedings of the 36th International Conference on Machine
  Learning}, 4013--4022.

\bibitem[{Liu and Tuzel(2016)}]{Liu2016CoupledGA}
Liu, M.-Y.; and Tuzel, O. 2016.
\newblock Coupled Generative Adversarial Networks.
\newblock In \emph{Advances in Neural Information Processing Systems},
  469--477.

\bibitem[{Liu et~al.(2019{\natexlab{b}})Liu, Ozay, Xu, Lin, and
  Okatani}]{generativedocking}
Liu, S.; Ozay, M.; Xu, H.; Lin, Y.; and Okatani, T. 2019{\natexlab{b}}.
\newblock A Generative Model of Underwater Images for Active Landmark Detection
  and Docking.
\newblock \emph{2019 IEEE/RSJ International Conference on Intelligent Robots
  and Systems (IROS)}, 8034--8039.

\bibitem[{Lloyd(1982)}]{kmeans}
Lloyd, S. 1982.
\newblock Least squares quantization in PCM.
\newblock \emph{IEEE transactions on information theory}, 28(2): 129--137.

\bibitem[{Luo et~al.(2019)Luo, Zheng, Guan, Yu, and Yang}]{Luo2019TakingAC}
Luo, Y.; Zheng, L.; Guan, T.; Yu, J.; and Yang, Y. 2019.
\newblock Taking a Closer Look at Domain Shift: Category-Level Adversaries for
  Semantics Consistent Domain Adaptation.
\newblock \emph{Proceedings of the IEEE Conference on Computer Vision and
  Pattern Recognition (CVPR)}, 2502--2511.

\bibitem[{Motiian et~al.(2017)Motiian, Jones, Iranmanesh, and
  Doretto}]{motiian2017few}
Motiian, S.; Jones, Q.; Iranmanesh, S.; and Doretto, G. 2017.
\newblock Few-shot adversarial domain adaptation.
\newblock In \emph{Advances in Neural Information Processing Systems
  (NeurIPS)}, 6670--6680.

\bibitem[{M{\"u}ller and Guido(2016)}]{muller2016introduction}
M{\"u}ller, A.~C.; and Guido, S. 2016.
\newblock \emph{Introduction to machine learning with Python: a guide for data
  scientists}.
\newblock " O'Reilly Media, Inc.".

\bibitem[{Peng, Wu, and Ernst(2018)}]{Peng2018ZeroShotDD}
Peng, K.-C.; Wu, Z.; and Ernst, J. 2018.
\newblock Zero-Shot Deep Domain Adaptation.
\newblock In Ferrari, V.; Hebert, M.; Sminchisescu, C.; and Weiss, Y., eds.,
  \emph{Proceedings of the European Conference on Computer Vision (ECCV)},
  793--810.

\bibitem[{Poggio, Banburski, and Liao(2020)}]{Poggio30039}
Poggio, T.; Banburski, A.; and Liao, Q. 2020.
\newblock Theoretical issues in deep networks.
\newblock \emph{Proceedings of the National Academy of Sciences}, 117(48):
  30039--30045.

\bibitem[{Rosenfeld, Ravikumar, and Risteski(2021)}]{rosenfeld2021the}
Rosenfeld, E.; Ravikumar, P.; and Risteski, A. 2021.
\newblock The Risks of Invariant Risk Minimization.
\newblock arXiv:2010.05761.

\bibitem[{Saxe, McClelland, and Ganguli(2019)}]{Saxe11537}
Saxe, A.~M.; McClelland, J.~L.; and Ganguli, S. 2019.
\newblock A mathematical theory of semantic development in deep neural
  networks.
\newblock \emph{Proceedings of the National Academy of Sciences}, 116(23):
  11537--11546.

\bibitem[{Sejnowski(2020)}]{Sejnowski30033}
Sejnowski, T.~J. 2020.
\newblock The unreasonable effectiveness of deep learning in artificial
  intelligence.
\newblock \emph{Proceedings of the National Academy of Sciences}, 117(48):
  30033--30038.

\bibitem[{Simonyan and Zisserman(2015)}]{VGG16}
Simonyan, K.; and Zisserman, A. 2015.
\newblock Very Deep Convolutional Networks for Large-Scale Image Recognition.
\newblock arXiv:1409.1556.

\bibitem[{Sun, Feng, and Saenko(2015)}]{sun2015return}
Sun, B.; Feng, J.; and Saenko, K. 2015.
\newblock Return of Frustratingly Easy Domain Adaptation.
\newblock arXiv:1511.05547.

\bibitem[{Sun and Saenko(2016)}]{sun2016deep}
Sun, B.; and Saenko, K. 2016.
\newblock Deep coral: Correlation alignment for deep domain adaptation.
\newblock In \emph{Proceedings of the European Conference on Computer Vision
  (ECCV)}, 443--450.

\bibitem[{Tang and Jia(2020)}]{Tang2020DiscriminativeAD}
Tang, H.; and Jia, K. 2020.
\newblock Discriminative Adversarial Domain Adaptation.
\newblock In \emph{Proceedings of the AAAI Conference on Artificial
  Intelligence}, 5940--5947.

\bibitem[{Tzeng et~al.(2017)Tzeng, Hoffman, Saenko, and
  Darrell}]{tzeng2017adversarial}
Tzeng, E.; Hoffman, J.; Saenko, K.; and Darrell, T. 2017.
\newblock Adversarial discriminative domain adaptation.
\newblock In \emph{Proceedings of the IEEE Conference on Computer Vision and
  Pattern Recognition (CVPR)}, 7167--7176.

\bibitem[{Tzeng et~al.(2014)Tzeng, Hoffman, Zhang, Saenko, and
  Darrell}]{tzeng2014deep}
Tzeng, E.; Hoffman, J.; Zhang, N.; Saenko, K.; and Darrell, T. 2014.
\newblock Deep Domain Confusion: Maximizing for Domain Invariance.
\newblock arXiv:1412.3474.

\bibitem[{van~der Maaten and Hinton(2008)}]{tsne}
van~der Maaten, L.; and Hinton, G. 2008.
\newblock Visualizing Data using t-SNE.
\newblock \emph{Journal of Machine Learning Research}, 9(86): 2579--2605.

\bibitem[{Wang, Cheng, and Jiang(2021)}]{DSPZDA}
Wang, J.; Cheng, M.-M.; and Jiang, J. 2021.
\newblock Domain Shift Preservation for Zero-Shot Domain Adaptation.
\newblock \emph{IEEE Transactions on Image Processing}, 30: 5505--5517.

\bibitem[{Wang and Jiang(2019)}]{Wang2019ConditionalCG}
Wang, J.; and Jiang, J. 2019.
\newblock Conditional Coupled Generative Adversarial Networks for Zero-Shot
  Domain Adaptation.
\newblock 3374--3383.

\bibitem[{Wang and Jiang(2020)}]{wang2020adversarial}
Wang, J.; and Jiang, J. 2020.
\newblock Adversarial Learning for Zero-shot Domain Adaptation.
\newblock In \emph{Proceedings of the European Conference on Computer Vision
  (ECCV)}, 329--344.

\bibitem[{Wang and Jiang(2021)}]{LATZDA}
Wang, J.; and Jiang, J. 2021.
\newblock Learning across Tasks for Zero-Shot Domain Adaptation from a Single
  Source Domain.
\newblock \emph{IEEE Transactions on Pattern Analysis and Machine
  Intelligence}.

\bibitem[{Wang et~al.(2019)Wang, Zhang, Yuan, and Feng}]{Wang2019FewShotAF}
Wang, T.; Zhang, X.; Yuan, L.; and Feng, J. 2019.
\newblock Few-Shot Adaptive Faster R-CNN.
\newblock \emph{Proceedings of the IEEE Conference on Computer Vision and
  Pattern Recognition (CVPR)}, 7166--7175.

\bibitem[{Xia and Ding(2020)}]{Xia2020HGNetHG}
Xia, H.; and Ding, Z. 2020.
\newblock HGNet: Hybrid Generative Network for Zero-Shot Domain Adaptation.
\newblock In \emph{Proceedings of the European Conference on Computer Vision
  (ECCV)}, 55--70.

\bibitem[{Xiao, Rasul, and Vollgraf(2017)}]{FashionMNIST}
Xiao, H.; Rasul, K.; and Vollgraf, R. 2017.
\newblock Fashion-MNIST: a Novel Image Dataset for Benchmarking Machine
  Learning Algorithms.
\newblock arXiv:1708.07747.

\bibitem[{Xu et~al.(2019{\natexlab{a}})Xu, Zhang, Ni, Li, Wang, Tian, and
  Zhang}]{xu2019adversarial}
Xu, M.; Zhang, J.; Ni, B.; Li, T.; Wang, C.; Tian, Q.; and Zhang, W.
  2019{\natexlab{a}}.
\newblock Adversarial Domain Adaptation with Domain Mixup.
\newblock arXiv:1912.01805.

\bibitem[{Xu et~al.(2019{\natexlab{b}})Xu, Zhou, Venkatesan, Swaminathan, and
  Majumder}]{xu2019d}
Xu, X.; Zhou, X.; Venkatesan, R.; Swaminathan, G.; and Majumder, O.
  2019{\natexlab{b}}.
\newblock d-SNE: Domain Adaptation Using Stochastic Neighborhood Embedding.
\newblock In \emph{Proceedings of the IEEE Conference on Computer Vision and
  Pattern Recognition (CVPR)}, 2492--2501.

\bibitem[{Zhang et~al.(2019)Zhang, Liu, Long, and Jordan}]{MDD_ICML_19}
Zhang, Y.; Liu, T.; Long, M.; and Jordan, M. 2019.
\newblock Bridging Theory and Algorithm for Domain Adaptation.
\newblock In \emph{International Conference on Machine Learning}, 7404--7413.

\bibitem[{Zhao et~al.(2019)Zhao, Combes, Zhang, and Gordon}]{zhao19a}
Zhao, H.; Combes, R. T.~D.; Zhang, K.; and Gordon, G. 2019.
\newblock On Learning Invariant Representations for Domain Adaptation.
\newblock In Chaudhuri, K.; and Salakhutdinov, R., eds., \emph{Proceedings of
  the 36th International Conference on Machine Learning}, volume~97 of
  \emph{Proceedings of Machine Learning Research}, 7523--7532.

\end{thebibliography}

\end{document}


\maketitle

\section{Qualitative Results for Feature Separability}
\begin{figure}
\centering
\subfigure[$p(x), a=D_E$]{
\includegraphics[width=0.31\textwidth]{pic/supple/tsne/s1_2-s2_3/domainout-tsne-mnist2mnistm-Emnist2Emnistm-compare-fromfile.png}
\label{subfig:domainout_dm_de}}
\subfigure[$p(x), a=D_F$]{
\includegraphics[width=0.31\textwidth]{pic/supple/tsne/s1_2-s2_3/domainout-tsne-mnist2mnistm-fmnist2fmnistm-compare-fromfile.png}
\label{subfig:domainout_dm_df}}
\subfigure[$p(x), a=D_N$]{
\includegraphics[width=0.31\textwidth]{pic/supple/tsne/s1_2-s2_3/domainout-tsne-mnist2mnistm-Nist2Nistm-compare-fromfile.png}
\label{subfig:domainout_dm_dn}}\\
\subfigure[$k(x), a=D_E$]{
\includegraphics[width=0.31\textwidth]{pic/supple/tsne/s1_2-s2_3/shareout-tsne-mnist2mnistm-Emnist2Emnistm-compare-fromfile.png}
\label{subfig:shareout_dm_de}}
\subfigure[$k(x), a=D_F$]{
\includegraphics[width=0.31\textwidth]{pic/supple/tsne/s1_2-s2_3/shareout-tsne-mnist2mnistm-fmnist2fmnistm-compare-fromfile.png}
\label{subfig:shareout_dm_df}}
\subfigure[$k(x), a=D_N$]{
\includegraphics[width=0.31\textwidth]{pic/supple/tsne/s1_2-s2_3/shareout-tsne-mnist2mnistm-Nist2Nistm-compare-fromfile.png}
\label{subfig:shareout_dm_dn}}\\
\subfigure[]{
\includegraphics[width=0.31\textwidth]{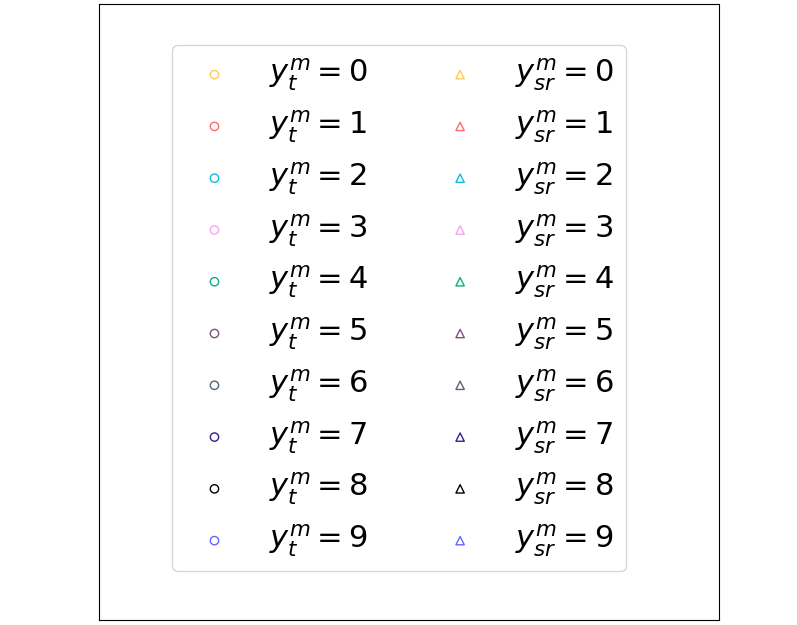}}
\caption{Visualization of distribution of feature representations using t-SNE when the domain pair is \textit{(G-dom,C-dom)} and main task is $D_M$. The first, second and third column gives features learned when the main task is $D_M$ while the auxiliary task is $D_E$, $D_F$ and $D_N$, respectively. Classes are indicated by different colors.  Triangle markers indicate features belonging to the source domain. Features belonging to the target domain are indicated by circle markers.}
\label{fig:dm}
\end{figure}

\begin{figure}
\centering
\subfigure[$p(x), a=D_E$]{
\includegraphics[width=0.31\textwidth]{pic/supple/tsne/s1_2-s2_3/domainout-tsne-fmnist2fmnistm-Emnist2Emnistm-compare-fromfile.png}
\label{subfig:domainout_df_de}}
\subfigure[$p(x), a=D_M$]{
\includegraphics[width=0.31\textwidth]{pic/supple/tsne/s1_2-s2_3/domainout-tsne-fmnist2fmnistm-mnist2mnistm-compare-fromfile.png}
\label{subfig:domainout_df_dm}}
\subfigure[$p(x), a=D_N$]{
\includegraphics[width=0.31\textwidth]{pic/supple/tsne/s1_2-s2_3/domainout-tsne-fmnist2fmnistm-Nist2Nistm-compare-fromfile.png}
\label{subfig:domainout_df_dn}}\\
\subfigure[$k(x), a=D_E$]{
\includegraphics[width=0.31\textwidth]{pic/supple/tsne/s1_2-s2_3/shareout-tsne-fmnist2fmnistm-Emnist2Emnistm-compare-fromfile.png}
\label{subfig:shareout_df_de}}
\subfigure[$k(x), a=D_M$]{
\includegraphics[width=0.31\textwidth]{pic/supple/tsne/s1_2-s2_3/shareout-tsne-fmnist2fmnistm-mnist2mnistm-compare-fromfile.png}
\label{subfig:shareout_df_dm}}
\subfigure[$k(x), a=D_N$]{
\includegraphics[width=0.31\textwidth]{pic/supple/tsne/s1_2-s2_3/shareout-tsne-fmnist2fmnistm-Nist2Nistm-compare-fromfile.png}
\label{subfig:shareout_df_dn}}
\subfigure[Legend]{
\includegraphics[width=0.31\textwidth]{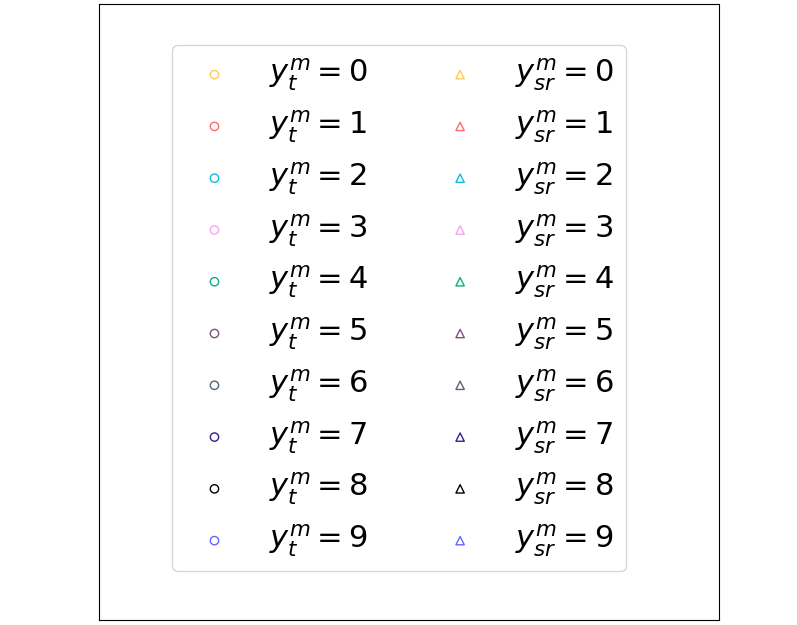}}
\caption{Visualization of distribution of feature representations using t-SNE when the domain pair is \textit{(G-dom,C-dom)} and main task is $D_F$. The first, second and third column gives features learned when the main task is $D_F$ while the auxiliary task is $D_E$, $D_M$ and $D_N$, respectively. Classes are indicated by different colors.  Triangle markers indicate features belonging to the source domain. Features belonging to the target domain are indicated by circle markers.}
\label{fig:df}
\end{figure}

\begin{figure}
\centering
\subfigure[$x_t^m, m=D_E$]{
\includegraphics[width=0.31\textwidth]{pic/supple/tsne/s1_2-s2_3/img-tsne-Emnist2Emnistm-mnist2mnistm-compare-fromfile.png}
\label{subfig:input_de}}
\subfigure[Legend]{
\includegraphics[width=0.31\textwidth]{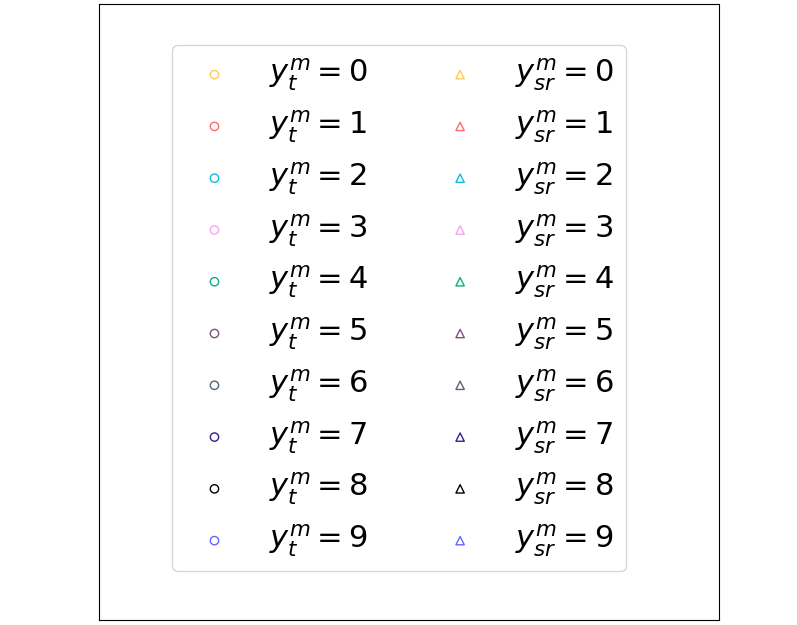}}\\
\subfigure[$p(x), a=D_M$]{
\includegraphics[width=0.31\textwidth]{pic/supple/tsne/s1_2-s2_3/domainout-tsne-Emnist2Emnistm-mnist2mnistm-compare-fromfile.png}
\label{subfig:domainout_de_dm}}
\subfigure[$k(x), a=D_M$]{
\includegraphics[width=0.31\textwidth]{pic/supple/tsne/s1_2-s2_3/shareout-tsne-Emnist2Emnistm-mnist2mnistm-compare-fromfile.png}
\label{subfig:shareout_de_dm}}
\subfigure[$q(x), a=D_M$]{
\includegraphics[width=0.31\textwidth]{pic/supple/tsne/s1_2-s2_3/fcin-tsne-Emnist2Emnistm-mnist2mnistm-compare-fromfile.png}
\label{subfig:fcin_de_dm}}\\

\subfigure[$p(x), a=D_F$]{
\includegraphics[width=0.31\textwidth]{pic/supple/tsne/s1_2-s2_3/domainout-tsne-Emnist2Emnistm-fmnist2fmnistm-compare-fromfile.png}
\label{subfig:domainout_de_df}}
\subfigure[$k(x), a=D_F$]{
\includegraphics[width=0.31\textwidth]{pic/supple/tsne/s1_2-s2_3/shareout-tsne-Emnist2Emnistm-fmnist2fmnistm-compare-fromfile.png}
\label{subfig:shareout_de_df}}
\subfigure[$q(x), a=D_F$]{
\includegraphics[width=0.31\textwidth]{pic/supple/tsne/s1_2-s2_3/fcin-tsne-Emnist2Emnistm-fmnist2fmnistm-compare-fromfile.png}
\label{subfig:fcin_de_df}}\\
\caption{Visualization of distribution of feature representations using t-SNE when the domain pair is \textit{(G-dom,C-dom)} and main task is $D_E$. The second and third row gives features learned when the main task is $D_E$ while the auxiliary task is $D_M$ and $D_F$, respectively. The first row shows $x_t^m$ where main task is $D_E$. Classes are indicated by different colors. Triangle markers indicate features belonging to the source domain. Features belonging to the target domain are indicated by circle markers.}
\label{fig:de}
\end{figure}

\begin{figure}
\centering
\subfigure[$x_t^m, m=D_N$]{
\includegraphics[width=0.31\textwidth]{pic/supple/tsne/s1_2-s2_3/img-tsne-Nist2Nistm-mnist2mnistm-compare-fromfile.png}
\label{subfig:input_dn_dm}}
\subfigure[Legend]{
\includegraphics[width=0.31\textwidth]{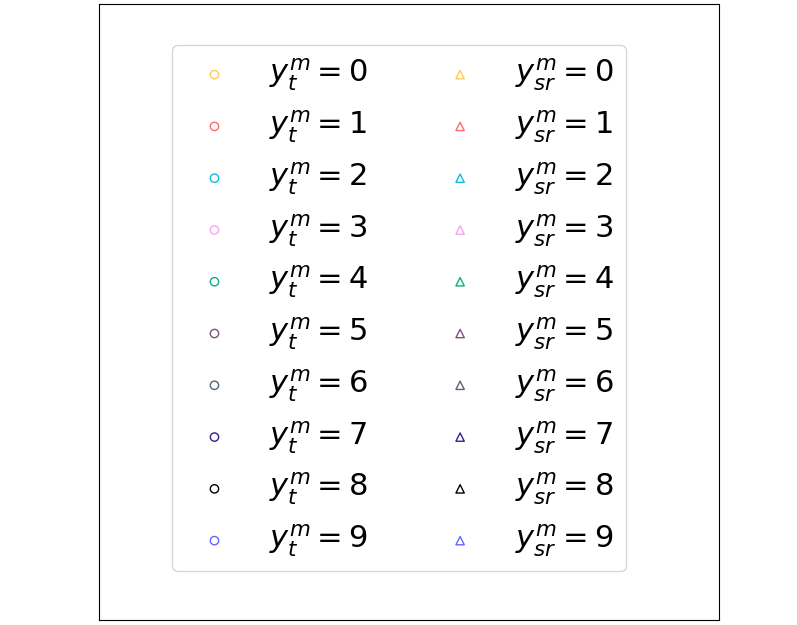}}\\
\subfigure[$p(x), a=D_M$]{
\includegraphics[width=0.31\textwidth]{pic/supple/tsne/s1_2-s2_3/domainout-tsne-Nist2Nistm-mnist2mnistm-compare-fromfile.png}
\label{subfig:domainout_dn_dm}}
\subfigure[$k(x), a=D_M$]{
\includegraphics[width=0.31\textwidth]{pic/supple/tsne/s1_2-s2_3/shareout-tsne-Nist2Nistm-mnist2mnistm-compare-fromfile.png}
\label{subfig:shareout_dn_dm}}
\subfigure[$q(x), a=D_M$]{
\includegraphics[width=0.31\textwidth]{pic/supple/tsne/s1_2-s2_3/fcin-tsne-Nist2Nistm-mnist2mnistm-compare-fromfile.png}
\label{subfig:fcin_dn_dm}}\\
\subfigure[$p(x), a=D_F$]{
\includegraphics[width=0.31\textwidth]{pic/supple/tsne/s1_2-s2_3/domainout-tsne-Nist2Nistm-fmnist2fmnistm-compare-fromfile.png}
\label{subfig:domainout_de_df}}
\subfigure[$k(x), a=D_F$]{
\includegraphics[width=0.31\textwidth]{pic/supple/tsne/s1_2-s2_3/shareout-tsne-Nist2Nistm-fmnist2fmnistm-compare-fromfile.png}
\label{subfig:shareout_de_df}}
\subfigure[$q(x), a=D_F$]{
\includegraphics[width=0.31\textwidth]{pic/supple/tsne/s1_2-s2_3/fcin-tsne-Nist2Nistm-fmnist2fmnistm-compare-fromfile.png}
\label{subfig:fcin_de_df}}\\
\caption{Visualization of distribution of feature representations using t-SNE when the domain pair is \textit{(G-dom,C-dom)} and main task is $D_N$. The second and third row gives features learned when the main task is $D_N$ while the auxiliary task is $D_M$ and $D_F$, respectively. The first row shows $x_t^m$ where main task is $D_N$. Classes are indicated by different colors. Triangle markers indicate features belonging to the source domain. Features belonging to the target domain are indicated by circle markers.}
\label{fig:dn}
\end{figure}

This section provides visualization of distributions of feature representations learned using the proposed TM-ZDA. Distributions of high dimensional features are visualized in two-dimensional spaces using t-SNE \cite{tsne}. The following distributions of feature representations are computed using \textit{(G-dom,C-dom)}  domain pair. Figure \ref{fig:dm} and \ref{fig:df} shows  distributions of feature representations when main task is $D_M$ and $D_F$, respectively. Feature representations of various segments in TD-ZDA are shown in Figure \ref{fig:de} and \ref{fig:dn} when the main task is $D_E$ and $D_N$, respectively. We define $p(x) \triangleq f_d^{t}(x)$, $k(x) \triangleq f_{s} \circ f_d^{t}(x)$ and $q(x) \triangleq f_{ta}^{m}  \circ f_{s} \circ f_d^{t}(x)$, where $x$ denotes either $x_t^{m}$ or $x_{sr}^{m}$. Triangle markers of the features depicted in the figures indicate that the corresponding features belong to the source domain. Features belonging to the target domain are indicated by circle markers.

\section{Implementation Details}
Adam algorithm \cite{Kingma2015AdamAM} is utilized to train TM-ZDA models with learning rate 0.0006. In the experiments, learning rate decay is not used, the batch size is $32$ and the total number of epochs is  $80$. We set the hyperparameter $\gamma$ to $50.0$. All the experiments are implemented using PyTorch \cite{pytorch}. The code used to reproduce the results will be provided in the camera ready version of the paper.
\bibliography{egbib}